\documentclass[10pt,reqno]{amsart}
\usepackage[foot]{amsaddr}
\usepackage{amssymb}
\usepackage{amsbsy}
\usepackage{amsthm}
\calclayout

\newcommand{\stepsize}{\ensuremath{h}}
\newcommand{\kineticenergy}{\ensuremath{T}}
\newcommand{\potentialenergy}{\ensuremath{V}}
\newcommand{\hamiltonian}{\ensuremath{H}}

\newtheorem{theorem}{Theorem}[section]

\newtheorem{example}{Example}[theorem]

\usepackage{glocalmacros}
\usepackage{amsmath}
\usepackage{graphicx}
\usepackage{tikz}
\usepackage[normalem]{ulem} 

\title[Structure preserving deep learning]{Structure preserving deep learning}

\makeatletter
\renewcommand{\email}[2][]{%
  \ifx\emails\@empty\relax\else{\g@addto@macro\emails{,\space}}\fi%
  \@ifnotempty{#1}{\g@addto@macro\emails{\textrm{(#1)}\space}}%
  \g@addto@macro\emails{#2}%
}
\makeatother
\author[Celledoni, Ehrhardt, Etmann, McLachlan, Owren, Sch\"onlieb and Sherry]{%
  Elena~Celledoni$\,^1$, Matthias~J.~Ehrhardt$\,^2$, Christian~Etmann$\,^3$, Robert~I.~McLachlan$\,^4$, Brynjulf~Owren$\,^1$, Carola-Bibiane~Sch\"onlieb$\,^3$ and Ferdia~Sherry$\,^3$}

\address{$^1$Department of Mathematical Sciences, NTNU, N-7491 Trondheim, Norway. Email addresses: \textup{\texttt{elena.celledoni@ntnu.no; brynjulf.owren@ntnu.no}}}
\address{$^2$Institute for Mathematical Innovation, University of Bath, Bath BA2 7JU, UK. Email address: \textup{\texttt{m.ehrhardt@bath.ac.uk}}}
\address{$^3$Department of Applied Mathematics and Theoretical Physics, University of Cambridge, Wilberforce Road, Cambridge CB3 0WA, UK. Email addresses: \textup{\texttt{cetmann@damtp.cam.ac.uk; cbs31@cam.ac.uk; fs436@cam.ac.uk}}}
\address{$^4$Institute of Fundamental Sciences, Massey University, Private Bag 11-222, Palmerston North, New Zealand. Email address: \textup{\texttt{r.mclachlan@massey.ac.nz}}}

\date{\today}

\begin{document}

\begin{abstract}
Over the past few years, deep learning has risen to the foreground as a topic of massive interest, mainly as a result of successes obtained in solving large-scale image processing tasks. There are multiple challenging mathematical problems involved in applying deep learning: most deep learning methods require the solution of hard optimisation problems, and a good understanding of the tradeoff between computational effort, amount of data and model complexity is required to successfully design a deep learning approach for a given problem. A large amount of progress made in deep learning has been based on heuristic explorations, but there is a growing effort to mathematically understand the structure in existing deep learning methods and to systematically design new deep learning methods to preserve certain types of structure in deep learning. In this article, we review a number of these directions: some deep neural networks can be understood as discretisations of dynamical systems, neural networks can be designed to have desirable properties such as invertibility or group equivariance, and new algorithmic frameworks based on conformal Hamiltonian systems and Riemannian manifolds to solve the optimisation problems have been proposed. We conclude our review of each of these topics by discussing some open problems that we consider to be interesting directions for future research.
\end{abstract}
\maketitle
\section{Introduction}
Structure preserving numerical schemes have their roots in geometric integration \cite{hairer2006geometric}, and numerical schemes that build on characterisations of PDEs as metric gradient flows \cite{ambrosio2008gradient}, just to name a few. The overarching aim of structure preserving numerics is to preserve certain properties of the continuous model, e.g mass or energy conservation, in its discretisation. But structure preservation is not just restricted to play a role in classical numerical analysis of ODEs and PDEs. Indeed, through the advent of continuum interpretations of neural networks \cite{haber2017stable, E2017deepode, E2018, Ruthotto2019pde}, structure preservation is also entering the field of deep learning. Here, the main objectives are to use the continuum model and structure preserving schemes to derive stable and converging neural networks and associated training procedures, and algorithms, e.g. neural networks which generalise well.

\subsection{Neural Networks}
Neural networks are a rich class of machine learning models that can be leveraged for many different tasks including regression, classification, natural language processing, reinforcement learning and image generation \cite{lecun2015deep}. While it is difficult to provide an all-encompassing definition for neural networks, they can generally be characterised as a combination of simple, parametric functions between \emph{feature spaces}. These functions act as individual building blocks (commonly called the \emph{layers} of the network). The main mechanism for combining these layers, which we will adopt in this work, is by function composition. 

For any $k \in \lbrace 0, \dots, K-1 \rbrace$, let $\FeatureSpace^k$ denote a vector space (our \emph{feature space}). While in most applications, these are simply finite-dimensional Euclidean spaces, we will assume more general structures (such as Banach spaces) when appropriate. With this, we then define a generic layer $$f^k: \FeatureSpace^{k} \times \GenParamSet^{k} \to \FeatureSpace^{k+1},$$
where $\GenParamSet^k$ is the set of possible parameter values of this layer. A neural network 
\begin{equation}
    \begin{aligned}
        \Psi: \FeatureSpace \times \GenParamSet &\to \LabelSpace \\
         (x,\theta) &\mapsto z^K
    \end{aligned}
\end{equation}
can then be defined via the iteration
\begin{equation}
\begin{aligned}
    \HiddenVar^0 &= x\\
    \HiddenVar^{k+1} &= f^k(\HiddenVar^{k},\theta^ k), \quad k = 0,\dots, K-1, \label{eq:neuralnet_general_form}
\end{aligned}
\end{equation}
such that $\FeatureSpace^0=\FeatureSpace$ and $\FeatureSpace^K=\LabelSpace$, where $\theta:=(\theta^0,\dots,\theta^{K-1}) \in \GenParamSet^0 \times \cdots \times \GenParamSet^{K-1}=:\Theta$ denotes the entirety of the network's parameters. The first layer is commonly referred to as the neural network's \emph{input layer}, the final layer as the neural network's \emph{output layer}, and all of the remaining layers are called \emph{hidden layers}.%\newline

While the above definitions are still quite general, in practice several standard layer types are employed. Most ubiquitous are those that can be written as a learnable affine combination of the input, followed by a simple, nonlinear function: the layer's \emph{activation function}. The quintessential example are \emph{fully-connected layers}
\begin{equation}
    \begin{aligned}
         f: \mathbb{R}^M \times (\mathbb{R}^{M'\times M} \times \mathbb{R}^{M'}) & \to \mathbb{R}^{M'} \\ 
         (\HiddenVar,(A,b)) &\mapsto \sigma ( A\HiddenVar + b),
    \end{aligned}
    \label{eq:fully_connected}
\end{equation}
whose parameters are the \emph{weight matrix} $A \in \mathbb{R}^{M'\times M}$ and the \emph{bias vector} $b \in \mathbb{R}^{M'}$. Its activation function $\sigma: \mathbb{R}^{M'} \to \mathbb{R}^{M'}$ is typically applied component-wise, e.g. the hyperbolic tangent $[\tanh (\HiddenVar)]_i:= \tanh (\HiddenVar_i)$ or the \emph{rectified linear unit} (ReLU) $[\text{relu}(\HiddenVar)]_i:=\max (0, \HiddenVar_i)$. For classification networks, the most common choice for the output layer's activation function is the \emph{softmax} activation function given by $$[\text{softmax}(z)]_i=\frac{\exp(z_i)}{ \sum^{M'}_{j=1} \exp (z_j) },$$%
%$[\text{softmax}(z)]_i = e^{z_i} / \sum_j e^{z_j}$%
such that the neural network's output's entries can be regarded as the individual class membership probabilities of the input. %\newline 
An important extension to the concept of fully-connected layers lie in \emph{convolutional layers} \cite{lecun1989backpropagation} where the matrix-vector product is replaced by the application of a (multi-channel) convolution operator. These are the main building block of neural networks used in imaging applications.

Suppose we are given a set of paired training data $(x_n, y_n)_{n=1}^N \subset \FeatureSpace \times \LabelSpace$, which is the case for predictive tasks like regression or classification. Training the model then amounts to solving the optimisation problem
\begin{align}
    \min_{\theta \in \Theta} \left\{ E(\theta) = \frac1N \sum_{n=1}^N \Loss_n (\Psi(x_n, \theta)) + R(\theta)\right\} . \label{eq1:training}
\end{align}
Here $L_n(y) := L(y, y_n) : \LabelSpace \to \mathbb R_\infty$ is the loss for a specific data point where $\Loss : \LabelSpace \times \LabelSpace \to \mathbb R_\infty := \mathbb R \cup \{\infty\}$ is a general loss function which usually satisfies $\Loss \geq 0$ and $\Loss(y_1, y_2) = 0$ if and only if $y_1 = y_2$. The function $L_n$ is usually smooth on its effective domain $\{y \mid L_n(y) < \infty\}$ and convex. $R : \GenParamSet \to \mathbb R_\infty$ acts as a regulariser which penalises and constrains unwanted solutions. In this setting, solving~\eqref{eq1:training} is a form of empirical risk minimisation \cite{shalev2014understanding}. Typically, variants of stochastic gradient descent are employed to solve this task. The calculation of the necessary gradients is performed using the famous \emph{backpropagation} algorithm, which can be understood as an application of reverse-mode auto-differentiation \cite{linnainmaa1970representation}.

\subsection{Residual Networks and Differential Equations}\label{sec:intro_resnets_ode}
In the following, we will discuss artificial neural networks architectures that arise from the numerical discretisation of time %dependent 
and parameter dependent differential equations. Differential equations have a long history in the mathematical treatment of neural networks. Initially, neural networks were motivated by biological neurons in the brain. Mathematical models for their interactions are based on nonlinear, time-dependent differential equations. These have inspired some famous artificial neural networks such as the Hopfield networks \cite{hopfield1982neural}. 

On a time interval $[0,T]$, the values of the approximation to the solution of the differential equation at different discrete times $0=t^0<t^1< \dots <t^K=T$, e.g. $t^k=k\,h$ and $h=T/K$%$h=\frac{T}{N}$
, corresponding to the different layers of the network architecture. For a fixed final time $T$, the existence of an underlying continuous model guarantees the existence of a continuous limit as the number of layers goes to infinity and $h$ goes to zero.

In the spirit of geometric numerical integration \cite{hairer2006geometric}, we discuss structural properties of the ANN as arising from the structure preserving discretisation of a differential equation with appropriate qualitative features such as having an underlying symmetry, an energy function or a Lyapunov function, preserving a volume form or a symplectic structure.

\begin{figure}
\def\PicWidth{3cm}
\newcommand{\PlotFrame}[4]{%
\node[anchor=south west] at (#4, 0) {\includegraphics[clip, trim=#2, width=\PicWidth]{#1}};%
\node[anchor=south west] at (#4, 0) {$k = #3$};%
}
\newcommand{\PlotFrames}[3]{%
\begin{tikzpicture}[x=\PicWidth+5mm, y=\PicWidth, inner sep=0pt]%
\PlotFrame{pics/#1_01}{#2}{0}{0}%
\PlotFrame{pics/#1_16}{#2}{15}{1}%
\PlotFrame{pics/#1_31}{#2}{30}{2}%
\PlotFrame{pics/#1_41}{#2}{40}{3}%
\node[anchor=west] at (0,0.5) {\rotatebox{90}{#3}};
\end{tikzpicture}%
}
\PlotFrames{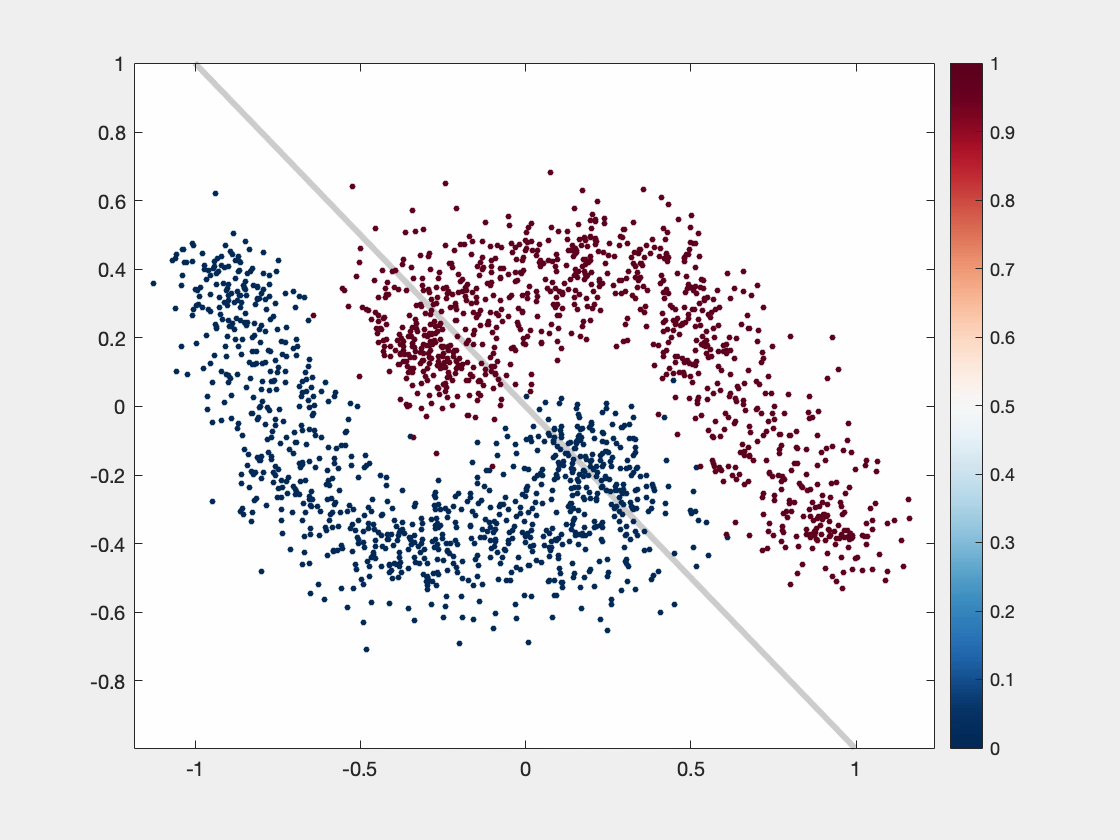}{140px 95px 190px 90px}{\texttt{halfmoon2d}}\\[3mm]
\PlotFrames{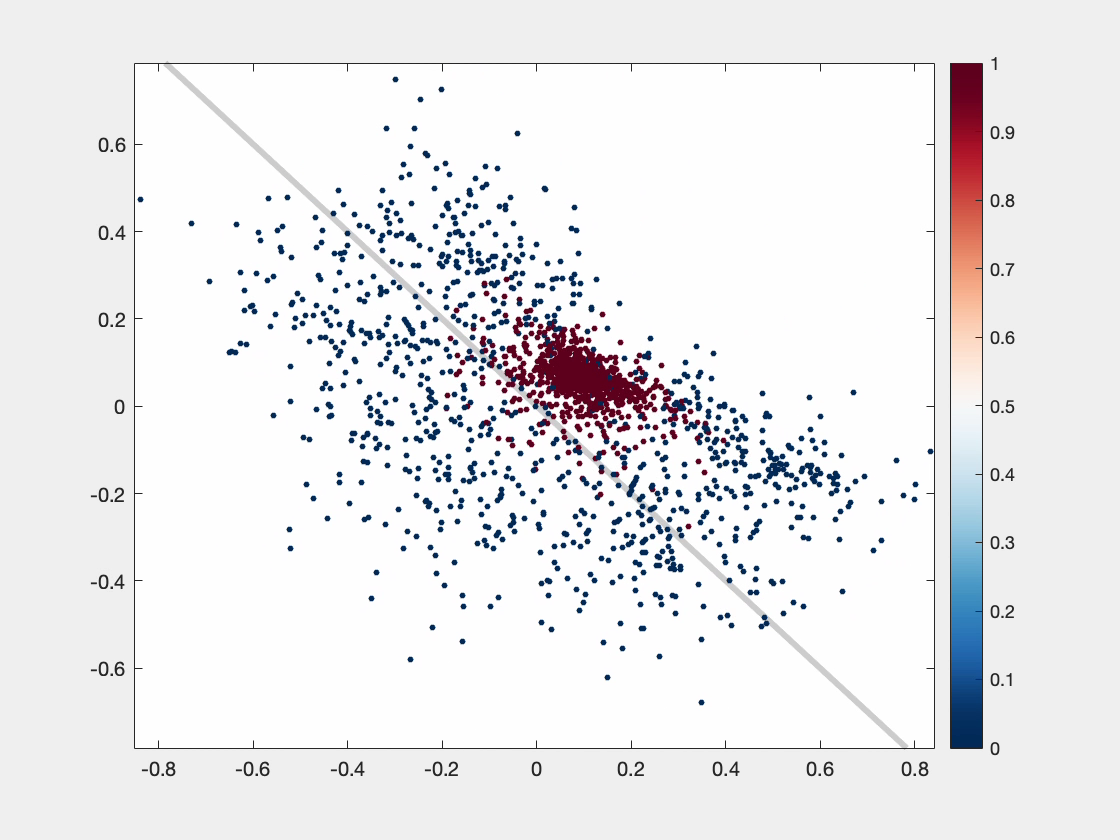}{140px 95px 195px 90px}{\texttt{donut2d}}\\[3mm]
\PlotFrames{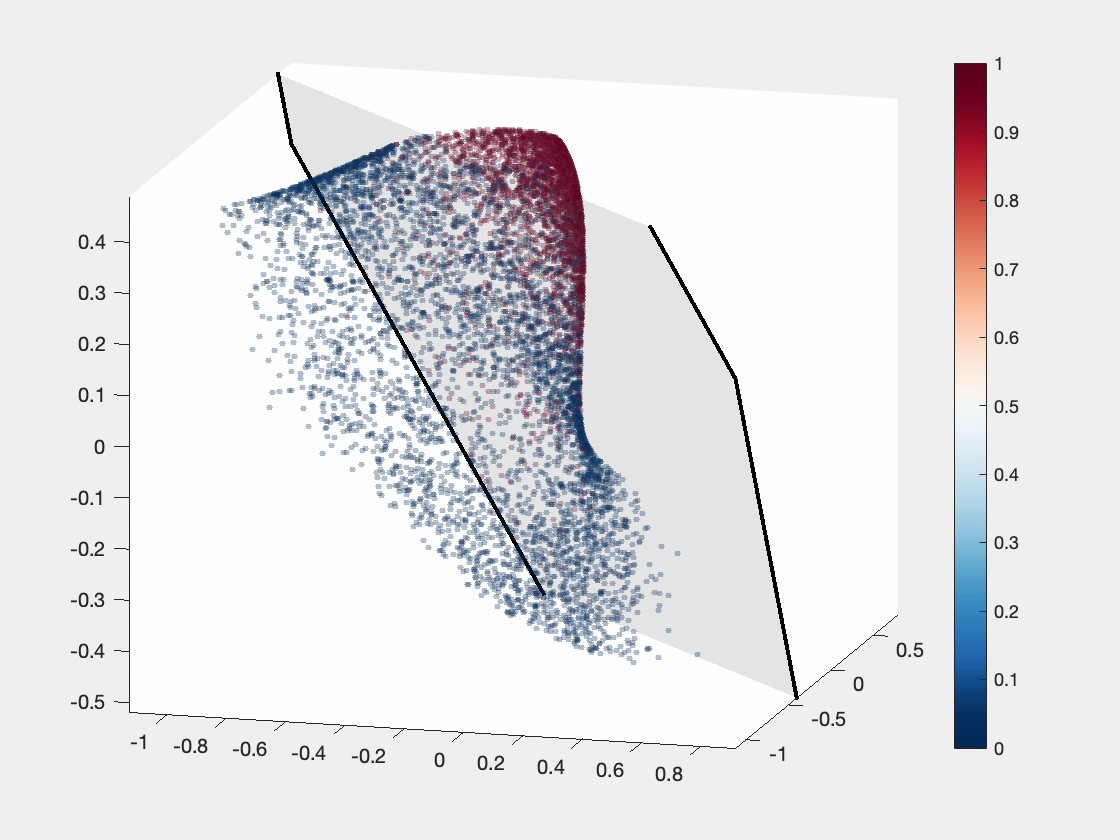}{210px 150px 350px 120px}{\texttt{donut3d}}
\caption{Evolutions of the ResNet model \eqref{eq1:discreterecursion2} for three different data sets. The upper two rows show evolutions in 2d and the lower row in 3d. The link function $\sigma$ is the hyperbolic tangent % $[\sigma(z)]_i = \tanh(z_i)$,
 and the data fit and regulariser are the squared 2-norm, $L_n(z) = \frac12 \|z-y_n\|^2_2$, $R(\theta) = \frac \lambda 2 \|\theta\|^2_2$.} \label{fig1:dynamics}
\end{figure}
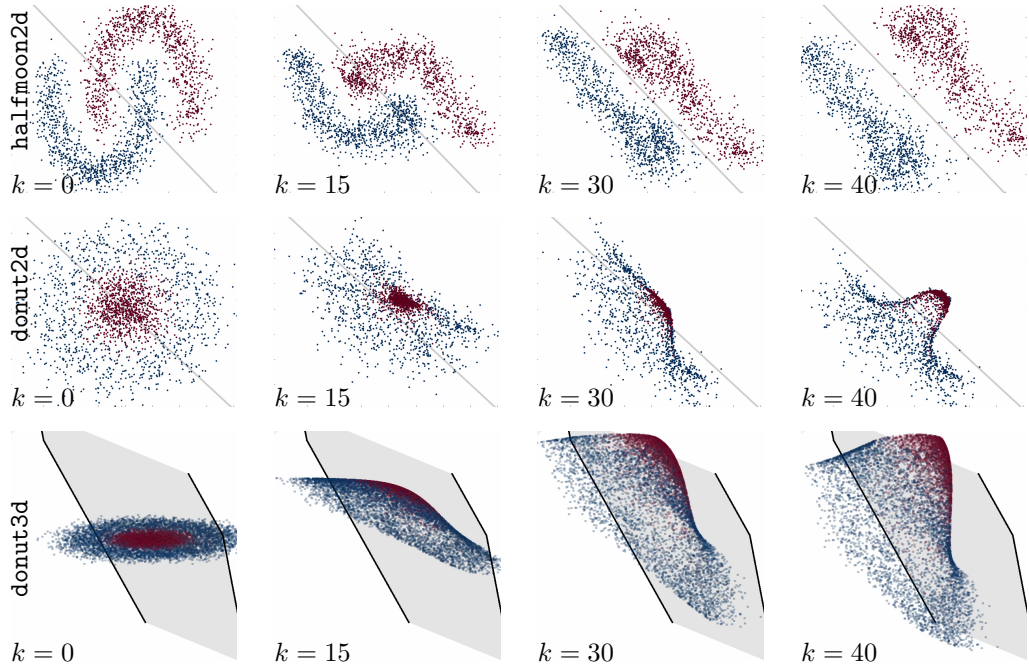

Contrary to the 'classical' design principle for layers of affine transformations followed by activation functions (cf. \eqref{eq:fully_connected}), so-called \emph{residual layers} are a variation to this principle, which has risen to become a standard design concept of neural networks. Here, the output of one such layer is again added to its input, which again defines a network, the ResNet~\cite{He2016resnet}, $\Psi: \FeatureSpace \times \GenParamSet \to \FeatureSpace, \Psi(x,\theta) = z^K$ that is now given by the iteration
\begin{equation}
\begin{aligned}
    \HiddenVar^0 &= x\\
    \HiddenVar^{k+1} &= \HiddenVar^{k} + \sigma(A^k \HiddenVar^{k} + b^k), \quad k = 0,\dots,K-1, \label{eq1:discreterecursion}
\end{aligned}
\end{equation}
if $\FeatureSpace=\LabelSpace$. If on the other hand the output space $\LabelSpace$ differs from $\FeatureSpace$, it is common to add another layer $\eta: \FeatureSpace \to \LabelSpace$ on top, which defines a network $\hat{\Psi}:=\eta \circ \Psi: \FeatureSpace \to \LabelSpace$. This is for example a common scenario in classification, where the dimensionality of $\LabelSpace$ is determined by the number of classes.

It is easy to see that \eqref{eq1:discreterecursion} corresponds to a particular discretisation of an ODE. To make the connection more precise, denote by %$t^k := k h, k = 0, \ldots, K, h = T/K$ a number of discrete points in $[0, T]$ and by
$z^k := \HiddenVar(t^k), A^k := A(t^k), b^k := b(t^k)$ samples of three functions $\HiddenVar : [0, T] \to \mathbb R^M, A : [0, T] \to \mathbb R^{M\times M}$ and $b : [0, T] \to \mathbb R^M$% at these locations
. With these notations we can write the ResNet \eqref{eq1:discreterecursion} as $\Psi(x, \theta) = \HiddenVar(T)$ with
\begin{align}
\HiddenVar(t^{k+1}) = \HiddenVar(t^k) + h \sigma(A(t^k) \HiddenVar(t^k) + b(t^k)), \quad k = 0, \ldots, K-1, \quad \HiddenVar(0) = x \label{eq1:discreterecursion2}
\end{align}
if $T = K$ and thus $h = 1$. For general $T$, it can be readily seen that \eqref{eq1:discreterecursion2} corresponds to the forward Euler discretisation of
\begin{align}
    \dot \HiddenVar(t) = f(\HiddenVar(t), \theta(t)), \quad t \in [0, T], \quad \HiddenVar(0) = x \label{eq1:ode}
\end{align}
with $\theta := (A, b) : [0, T] \to \mathbb R^{M\times M} \times \mathbb R^M\cong \mathbb R^{M^2+M}$,
\begin{equation} \label{eq1:vectorfield}
f(\HiddenVar(t), \theta(t)) = \sigma(A(t)\HiddenVar(t) + b(t)).
\end{equation}
Thus, there is a natural connection of the discrete deep learning problem \eqref{eq1:training}+\eqref{eq1:discreterecursion} with the optimal control formulation \eqref{eq1:training}+\eqref{eq1:ode}. The dynamics of the ResNet \eqref{eq1:discreterecursion} as a discretisation of \eqref{eq1:ode} is depicted in Figure \ref{fig1:dynamics}.

From here on we will suppress the dependence on $t$ whenever it is clear from the context.

It is a well-known fact that the optimal control formulation can be phrased as a closed dynamical system by using Pontryagin's principle and this results in a constrained Hamiltonian boundary value problem \cite{Benning2019ode}, the Hamiltonian of this system is given as
\begin{align}
    H(\HiddenVar, \LagrangeMultiplier, \genparam) = \langle\LagrangeMultiplier, f(\HiddenVar, \genparam)\rangle
    \label{eq1:hamiltonian}
\end{align}
where $\LagrangeMultiplier\in T^*\FeatureSpace\equiv\mathbb{R}^M$ is a vector of Lagrange multipliers. The dynamical system is then given by 
\begin{align}
\dot \HiddenVar = \partial_\LagrangeMultiplier H, \quad
\dot \LagrangeMultiplier = -\partial_\HiddenVar H, \label{eq1:extended_ode}
\end{align}
subject to the constraint
\begin{align}
 \quad
0 = \partial_\genparam H.  \label{eq1:constraint}
\end{align}
The adjoint equation for $\LagrangeMultiplier$ can be expressed as
\begin{align}
    \dot \LagrangeMultiplier = \partial_z f^T \LagrangeMultiplier, \quad
    p(T) &= \frac{1}{N}\sum_{n=1}^N \partial_\HiddenVar \Loss_n^\prime(z(T)).
\end{align}

In what follows we will review some of the guiding principles of structure preserving deep learning, and in particular recent contributions for new neural networks architectures as discretisations for ODEs and PDEs in Section~\ref{sec:ODEPDE} and the interpretation of the training of neural networks as an optimal control problem in Section~\ref{sec:optimalcontrol}, invertible neural networks in Section \ref{sec:invertible}, equivariant neural networks in Section \ref{sec:equivariant}, and structure-preserving numerical schemes for the training of neural networks in Section \ref{sec:SPT}.

\section{Neural networks inspired by differential equations}\label{sec:ODEPDE}
\subsection{Structure preserving ODE formulations} \label{sec:structure_preserving_ODE_formulations}
    In this section, we look at how the ODE formulation \eqref{eq1:ode} can be restricted or extended in order to ensure that its flow has favourable properties. We are zooming in on the forward problem itself, assuming that the parameters $\genparam$ are fixed. 
    By abuse of notaton, we will in this section write $f(t,z)$ for $f(z,\theta(t))$ in that we consider $\theta(t)$ a known function of $t$.
    It is perhaps not so obvious what the desirable properties of the flow should be, and to some extent we here lean on prior work by Haber and Ruthotto \cite{haber2017stable} as well as Chang et al. \cite{chang2018reversible}.
    It seems desirable that small perturbations in the data should not lead to large differences in the end result.  Preferably, the forward model should have good stability properties, for instance in the sense that for any two nearby solutions $z_1(t)$ and $z_2(t)$ to \eqref{eq1:ode} 
    \begin{equation} \label{endbound}
         \|z_2(T)-z_1(T)\| \leq  C \|z_2(0)-z_1(0)\|
    \end{equation}
    for a moderately sized constant $C$. It is well-known that this type of estimate can be obtained in several different ways, depending on the properties of the underlying vector field in \eqref{eq1:ode}. If $f(t,z)$ is Lipschitz in its second argument with constant $L$, then \eqref{endbound} holds with $C=e^{TL}$. Looking at \eqref{eq1:vectorfield}, one can use
    $L=L_\sigma\, \displaystyle{\max_t}\|A(t)\|$ where $L_\sigma$ is a Lipschitz constant for the activation function $\sigma$.

    Stability can also be studied in terms of Lyapunov functions, that is, functions $V(z)$ that are non-increasing along solution trajectories. Functions that are constant along solutions are called first integrals, and a particular instance is the energy function of autonomous Hamiltonian systems.
    
    For stability of nonlinear ODEs one may for instance consider growth and contractivity in the $L^2$-norm using a one-sided Lipschitz condition. This is similar to the analysis proposed in \cite{Zhang2020resnet}.
      We assume that there exists a constant $\nu\in\mathbb{R}$ such that for all admissible $z_1, z_2$, and $t\in[0,T]$ we have
      \begin{equation} \label{onesidedLip}
          \left\langle
      f(t,z_2)-f(t,z_1), z_2-z_1
      \right\rangle \leq \nu \|z_2-z_1\|^2.
      \end{equation}
 In this case it is easily seen that for any two solutions $z_1(t)$, $z_2(t)$ to \eqref{eq1:ode} one has  
 $$
    \|z_2(t)-z_1(t)\| \leq \|z_2(0)-z_1(0)\|\, e^{\nu t},
 $$
 so that the problem is contractive if $\nu \leq 0$ \cite{hairer2010ode2}.
 For instance, the vector field \eqref{eq1:vectorfield} satisfies \eqref{onesidedLip} for $\sigma$ absolutely continuous if
 $$
 \nu \leq\sup_{t,D}\, \lambda_{\max}\left( \frac{D A(t)+(D A(t))^T}{2}  \right),
 $$
 where the supremum is taken over all diagonal matrices $D$ with diagonal entries in $\sigma'(\mathbb{R})$.
 Some care should be taken here: It is {\em not} so that the sign of the eigenvalues of $\tfrac12 (DA(t)+(DA(t))^T)$ is invariant under the set of all positive diagonal matrices $D$. In particular, even if $A(t)$ is skew-symmetric, the vector field \eqref{eq1:vectorfield} may still have a positive one sided Lipschitz constant $\nu$, but if we assume for instance that $\sigma'(s)\in[0,1]$, it holds that $\nu\leq\frac12\,\displaystyle{\max_t}\, \|A(t)\|_\infty$.
 Haber and Ruthotto \cite{haber2017stable} suggest to use the eigenvalues of the linearised ODE vector field to analyse the stability behaviour of the forward problem. If the eigenvalues of the  resulting Jacobian matrix have only non-positive real parts, it may be an indication of stability, yet for non-autonomous systems such an analysis may lead to wrong conclusions.

 Clearly, the set of all vector fields of the form \eqref{eq1:vectorfield} include both stable and unstable cases. 
 There are different ways of ensuring that a vector field has stable or contractive trajectories. One could be to put restrictions on the family of matrices $A(t)$, another option is to alter the form of \eqref{eq1:vectorfield} either by adding symmetry to it or by embedding it into a larger space where it can be given desirable geometric properties, e.g. by doubling the dimension, it is possible in several different ways to obtain a non-autonomous Hamiltonian vector field. In \cite{chang2018reversible} and \cite{haber2017stable} several models are suggested, and we mention a few of them in Section~\ref{subsubsec:hamilton}.

 \subsubsection{Dissipative models and gradient flows.}
 
 For different reasons it is desirable to use vector fields with good stability properties. This will ensure that data which are close to each other in a chosen norm initially remain close as the features propagate through the neural network.
 A model suggested in \cite{haber2017stable} was to consider weight matrices $A(t)$ of the form
$$
    A(t) = S(t)-S(t)^T - \gamma I
$$
where $S(t)$ is arbitrary and $\gamma$ is a small dissipation parameter. Flows of vector fields of the form $\dot{z}=A(t)z+b(t)$ exactly preserve the $L^2$-norm of the flow when $A(t)$ is skew-symmetric. The non-linearity will generally alter this behaviour,  but adding a small amount of dissipation can improve the stability. It is however not guaranteed to reduce the one-sided Lipschitz condition.

Zhang and Schaeffer \cite{Zhang2020resnet} analyse the stability of a model where the activation function $\sigma$ is always chosen to be the {\em Rectified Linear Unit} (ReLU) function. The form of the discretised model is such that in the limit when $\stepsize$ tends to zero it must be replaced by a differential inclusion rather than an ODE of the form discussed above, meaning that $\dot z-f(t,z)$ belongs to some specified subdomain of $\mathbb{R}^M$ and their model vector field is
 \begin{equation} \label{eq3:zhangform}
   f(t,z) = -A_2(t)\sigma(A_1(t)z(t)+b_1(t))+b_2(t),
 \end{equation}
 Growth and stability estimates are derived for this class of vector fields as well as for cases where restrictions are imposed on the parameters, such as $A_2(t)$ having positive elements or  the case $A_1(t)=A_2(t)^T=:A(t)$ and $b_2(t)=0$. For this last case, we consider for simplicity the ODE
 \begin{equation} \label{gradsys}
     \dot{z} = -A(t)^T \sigma(A(t) z + b(t)) = f(t,z),
 \end{equation}
 which is a gradient system in the sense that $\dot{z}=-\nabla_z V$ with $V=\gamma(A(t)z+b(t))\mathbf{1}$ where $\gamma'=\sigma$ and
 $\mathbf{1}$ is the vector of ones. 
 \begin{theorem} \label{theo:gradsys}\hspace{0pt}\newline
 \begin{enumerate}
     \item  Let $V(t,z)$ be twice differentiable and convex in the second argument. Then the gradient vector field
 $f(t,z)=-\nabla V(t,z)$ satisfies a one-sided Lipschitz condition \eqref{onesidedLip} with $\nu\leq 0$.
 \item
 Suppose that $\sigma(s)$ is absolutely continuous and $0\leq\sigma'(s)\leq 1$ a.e. in $\mathbb{R}$. Then \eqref{gradsys} satisfies the one-sided Lipschitz condition \eqref{onesidedLip} for any choice of parameters $A(t)$ and $b(t)$ with 
 $$
     -\mu_*^2 \leq \nu_\sigma \leq 0 
 $$
 where $\mu_*=\displaystyle{\min_t}\, \mu(t)$ and where $\mu(t)$ is the smallest singular value of $A(t)$.
 In particular $\nu_\sigma=-\mu_*^2$ is obtained when $\sigma(s)=s$.
 \end{enumerate}
 \end{theorem}
 {\em Proof.} (1) We compute
 $$
 \langle f(t,z_2)-f(t,z_1),z_2-z_1\rangle = -\langle \nabla_zV(t,z_2)-\nabla_z V(t,z_1),z_2-z_1\rangle
 $$
 and define
 $$
 \phi(\xi) = \frac{\ddiff}{\ddiff\xi} V(t,\xi z_2 + (1-\xi)z_1)
 $$
 such that
 $$
 \langle \nabla_zV(t,z_2)-\nabla_z V(t,z_1),z_2-z_1\rangle=\phi(1)-\phi(0)=\int_0^1\phi'(\xi)\dint \xi.
 $$
 Therefore, by the convexity of $V(t,z)$
 $$
 \langle f(t,z_2)-f(t,z_1),z_2-z_1\rangle = -\langle \int_0^1\nabla_z^2 V(t,\xi z_2+(1-\xi) z_1) \dint\xi\; (z_2-z_1) , z_2-z_1 \rangle \leq 0
 $$
 (2) Let $z_1$ and $z_2$ be vectors in $\mathbb{R}^M$. Using \eqref{gradsys} while suppressing the $t$-dependence in the parameters, we find
 \begin{equation} \label{Proof21-1}
    \langle f(t,z_2)-f(t,z_1), z_2-z_1\rangle 
 = -\langle \sigma(A z_2 + b)- \sigma(A z_1 + b), A(z_2-z_1)\rangle 
 \end{equation}
 
 For real scalars $\zeta, \eta$ and $\beta$ we have
 $$
 (\sigma(\zeta+\beta)-\sigma(\eta+\beta))(\zeta-\eta)=\int_\eta^\zeta\sigma'(s+\beta)\, \dint s\, (\zeta-\eta)
 $$
 and since $0\leq \sigma'(s+\beta)\leq 1$ a.e. we have 
 $$0\leq (\sigma(\zeta+\beta)-\sigma(\eta+\beta))(\zeta-\eta)\leq (\zeta-\eta)^2.$$
 Using this inequality in \eqref{Proof21-1} we obtain
 $$
 -\|Az_2-Az_1\|^2 \leq  \langle f(t,z_2)-f(t,z_1), z_2-z_1\rangle  \leq 0
 $$
 Since $\|A(z_2-z_1)\|^2 \geq \mu_*^2\,\|z_2-z_1\|^2$ the result follows. $\Box$
 
 {\em Remark.} In Theorem~\ref{theo:gradsys} we restricted the class of activation functions to be absolutely continuous with 
 $0\leq \sigma'(s)\leq 1$. This is true for many of the activation functions proposed in the literature, in particular for the
 ReLU function and the sigmoid $\sigma(s)=\tanh s$.

\subsubsection{Hamiltonian vector fields.} \label{subsubsec:hamilton}

One may take inspiration from mechanical systems and introduce the Hamiltonian framework. 
Separable Hamiltonian systems are defined in terms of  kinetic and potential energy functions $\kineticenergy$ and $\potentialenergy$ 
$$
\hamiltonian(t,z,p) = \kineticenergy(t,p) + \potentialenergy(t,z)
$$
The leads to differential equations of the form

\begin{align}
    \dot{z} &= \partial_p \hamiltonian = \partial_p \kineticenergy \label{zdot}\\
    \dot{p} &= -\partial_z \hamiltonian = -\partial_z \potentialenergy \label{pdot}
\end{align}

There are different ways to construct a Hamiltonian system from \eqref{eq1:vectorfield}. 
In \cite{chang2018reversible} the following model is suggested
\begin{align*}
    \dot{z} &= A_1(t)^T \sigma_1(A_1(t) p + b_1(t)) \\
    \dot{p} &= -A_2(t)^T \sigma_2(A_2(t)z + b_2(t))
\end{align*}
Let $\gamma_i:\mathbb{R}\rightarrow\mathbb{R}$ be such that $\gamma_i'(t)=\sigma_i(t),\;i=1,2$. The corresponding Hamiltonian is
$$
    \kineticenergy(t,p) = \gamma_1(A_1(t)p+b_1(t))\mathbf{1},\quad \potentialenergy(t,z) = \gamma_2(A_2(t)z+b_2(t))\mathbf{1}
$$
 where $\mathbf{1}=(1,\ldots,1)^T$. 
 A simple case is obtained by choosing $\sigma_1(s):=s$, $A_1(t)\equiv I$, $b_1(t)\equiv 0$ and $\sigma_2(s):=\sigma(s)$ which after eliminating $p$ yields the second order ODE
 $$
       \ddot{z} = -\partial_z V =-A_2(t)^T \sigma(A_2(t)z + b_2(t))
 $$
 A second example considered in \cite{chang2018reversible} is obtained by setting $\sigma_1=\sigma_2=\sigma$.
 
 From the outset, it might not be so clear which geometric features such a non-autonomous Hamiltonian system has. There seem to be at least two ways to understand this problem \cite{marthinsen16gio}. Let us assume that $u = ( z, p)\in T^*\mathbb{R}^M\equiv \mathbb{R}^M\oplus\mathbb{R}^M$ with  ``positions" $z$ and  ``momenta" $p$ forming the phase space. We have the natural symplectic form on the phase space $\omega_0=\mathrm d p \wedge \mathrm d z$. 
This form can be represented by the Darboux-matrix $J$ as
$$
\omega_0(\xi,\eta) = \langle \xi, J\eta\rangle,\quad J=\left[\begin{array}{rr}0 & I \\ -I & 0\end{array}\right]
$$
and the Hamiltonian vector field $f(t,z,p)$ is defined via $\mathrm{d}\hamiltonian(\cdot)=\omega_0(f(t,z,p),\cdot)$.

Let us now introduce as phase space $T^*\mathbb{R}^M\times\mathbb{R}$ by including the time $t$ as a dependent variable. The natural projection is $\tau: (z,p,t)\mapsto (z,p)$. The space $T^*\mathbb{R}^M\times\mathbb{R}$ can then be furnished with a \emph{contact structure} defined as
$$
  \omega_\hamiltonian =  \tau^*\omega_0 - \mathrm{d}\hamiltonian \wedge \mathrm{d}t.
$$
The first term is just the original form,  ignoring the $t$-component of the tangents, the second form is only concerned with the ``$t$-direction". One can write the matrix of $\omega_\hamiltonian$  by adding a row and a column to $J$ coming from the second term 
$ - \mathrm{d\hamiltonian} \wedge \mathrm{d}t$
$$
     J_H=\left(
     \begin{array}{c|c}
     J & -\nabla_{u}\hamiltonian \\[3mm] \hline
     (\nabla_{u}\hamiltonian)^T & 0
     \end{array}
     \right)
$$
The resulting vector field is then $f_\hamiltonian=(f^T,f_t)^T$ and can be expressed through the equations
$$
      i_{f_\hamiltonian}\omega_H = 0,\quad i_{f_\hamiltonian} \mathrm{d}t=1
$$
where $\mathrm{i}_{f_\hamiltonian}$ stands for the interior derivative of the form by $f_\hamiltonian$, e.g. $\mathrm{i}_{f_\hamiltonian}$ applied to the two-form $\omega_\hamiltonian$ is the one-form $\alpha=\mathrm{i}_{f_\hamiltonian}\omega_\hamiltonian$ such that
$\alpha(\eta)=\omega(f_\hamiltonian,\eta)$ for all vector fields $\eta$.
The form $\omega_\hamiltonian$ is preserved along the flow, but $\hamiltonian$ is not.

\smallskip\noindent
\emph{Extended phase space.} One can in fact recover an autonomous Hamiltonian problem from the non-autonomous one by adding two extra dependent variables, say $t$ and $p_t$. We do this by considering the manifold $T^*(\mathbb{R}^M\times\mathbb{R})$ which can be identified with $T^*\mathbb{R}^{M+1}$. One needs a new projection
$\mu: (z,p,t,p_t)\mapsto (z,p,t)$ and we can define an extended (autonomous) Hamiltonian on $T^*(\mathbb{R}^M\times\mathbb{R})$ as
$$
      K = \hamiltonian\circ\mu + p_t
$$
with corresponding two-form
$$
      \Omega_0 = \mu^*\tau^*\omega_0 + \mathrm{d}p_t \wedge \mathrm{d}t
$$
The corresponding matrix, $J_E$, is just the original Darboux-matrix $J$ where each of the $n\times n$ identity matrices has been replaced by corresponding $(M+1)\times (M+1)$ identity matrices.
The extended Hamiltonian $K$ is exactly preserved along the flow so that
the new conjugate momentum variable $p_t$ will keep track of how $\hamiltonian(z,p,t)$ is varying along solutions. The resulting ODE vector feld $f_E$ can be written as $dK(\cdot)=\Omega_0(f_E,\cdot)$ and in coordinates the ODE system becomes
\begin{equation}\label{extendedHam}
\dot{z}=\partial_p \hamiltonian,\quad \dot{p}=-\partial_z \hamiltonian,\quad \dot{t}=1,\quad \dot p_t = -\partial_t \hamiltonian.
\end{equation}
We see at once that since the equations for $\dot{z}, \dot{p}$ do not depend on $p_t$ and since the solution for $t$ is explicitly given, we solve the same problem as before. After solving for $z$ and $p$, we obtain $p_t$ independently by integration. 
The second thing one may observe is that if a numerical method $\phi_{\stepsize}$ has the property that\footnote{It is common to assume that a given numerical integrator is defined on systems in any dimension}
$$
      \phi_{\stepsize} \circ \mu = \mu\circ \phi_{\stepsize}
$$
then what we obtain by just considering the first $2M$ components of the numerical solution to the extended system is precisely the same as what we would have obtained applying the same method to the original non-autonomous problem. This observation was used by Asorey et al. \cite{asorey83gct} to define what is meant by canonical transformations in the non-autonomous case, and we refer to this paper for further results on the structural connections between the two systems.
In applications to deep learning, one should note that geometric properties of the solution can mostly be deduced from the extended system rather than the original non-autonomous one, there are numerical methods which preserve energy or the symplectic form $\Omega_0$ and rigorous results can be proved for the long time behaviour of such integrators \cite{hairer2006geometric}.

\subsection{Structure preserving numerical methods for the ODE model} 
The rationale behind proposing ODE formulations with geometric structure is to enforce a controlled behaviour of the solution as it is propagated through the hidden layers of the network. It is therefore also important that when the ODE is approximated by a numerical method, this behaviour should be retained by the numerical scheme.

\subsubsection{Numerical methods preserving dissipativity}
When solving numerically ODEs which satisfy the one-sided Lipschitz condition \eqref{onesidedLip} a desirable property of the numerical scheme would be that it contracts two nearby numerical solutions whenever $\nu\leq 0$. That is, it should satisfy
\begin{equation}\label{eq:contractive}
     \|z_1^{k+1} - z_2^{k+1}\| \leq \|z_1^{k} - z_2^{k}\|,
\end{equation}
in each time step $k$, preferably without too severe restrictions on the time step $\stepsize$. Methods which can achieve this for any step size exist, and are called B-stable. There are many B-stable methods, and for Runge--Kutta methods B-stability can be easily checked by the condition of algebraic stability. Examples of B-stable methods are
the implicit Euler method, and all implicit Runge--Kutta methods within the classes, Gauss, Radau IA, Radau IIA and Lobatto IIIC \cite{hairer2010ode2}. In deep learning algorithms it has been more usual to consider explicit schemes since they are much cheaper per step than implicit ones, but are subject to restrictions on the time step used. Note for instance that the explicit Euler method is unstable for any step size when applied even to linear constant coefficient systems where there are eigenvalues of the coefficient matrix on the imaginary axis.
To consider contractivity of explicit schemes, we need to replace \eqref{onesidedLip} by a different monotonicity condition
$$
\langle f(t,z_2)-f(t,z_2), z_2-z_1  \rangle \leq \bar{\nu} \| f(t,z_2)-f(t,z_1)  \|^2
$$
where we assume $\bar{\nu}<0$. For every Runge--Kutta method with strictly positive weights $b_i$ there is a constant $r$ such that the numerical solution is contractive whenever the step size satisfies
$$
     \stepsize < -2\bar{\nu} r
$$
The value of $r$ can be calculated for every Runge--Kutta method with positive weights, and e.g. for the explicit Euler method as well as for the classical 4th order method of Kutta one has $r=1$ \cite{dahlquist79gdo}.

\subsubsection{Numerical methods preserving energy or symplecticity}
For autonomous Hamiltonian systems there are two important geometric properties which are conserved. One is the energy or Hamiltonian $\hamiltonian(z,p)$ the other one is the symplectic structure, a closed non-degenerate differential two-form. These two properties are also the main targets for structure preserving numerical methods.

All Hamiltonian deep learning models presented here can be extended to a separable, autonomous canonical system, i.e. a system of the form \eqref{zdot}-\eqref{pdot}. Such systems preserve the symplectic two-form $\mathrm{d}p\wedge \mathrm{d}q$ and there are many examples of explicit numerical methods that also preserve this same form in the sense that $\mathrm{d}p^{k+1}\wedge \mathrm{d}q^{k+1}= \mathrm{d}p^k\wedge \mathrm{d}q^k$. The simplest example of such a scheme is the symplectic Euler method, defined for the variables $z$ and $p$ as
\begin{equation}
    \label{sympEul}
  z^{k+1} = z^k + \stepsize\, \partial_p T(t^{k+1},p^k),\quad
  p^{k+1} = p^k - \stepsize\, \partial_z V (t^{k+1},z^{k+1})
  \end{equation}
%Another popular example is the St{\"o}rmer-Verlet integrator. Both
The symplectic Euler method is explicit for separable Hamiltonian systems and is an example of a splitting method \cite{mclachlan02sm}. 
Many other examples and a comprehensive treatment of symplectic integrators can be found in \cite{hairer2006geometric}.
When applying symplectic integrators to Hamiltonian problems one has the important notion of backward error analysis. The numerical approximations obtained can be interpreted as the exact flow of a perturbed system with Hamiltonian
$\tilde{\hamiltonian}(z,p)=\hamiltonian(z,p) + \stepsize H_2(z,p)+\cdots$\footnote{This is a divergent asymptotic series, but truncation is possible at the expense of an exponentially small remainder term}. This partly explains the popularity of symplectic integrators, since many of the characteristics of Hamiltonian flows are inherited by the numerical integrator. 

There exist many numerical methods which preserve energy exactly for autonomous problems, for instance there is a large class of schemes based on discrete gradients \cite{mclachlan99giu}. A discrete gradient of a function $\hamiltonian(z)$ is a continuous map $\overline{\nabla}\hamiltonian:\mathbb{R}^M\times\mathbb{R}^M\rightarrow\mathbb{R}^M$ which satisfies the following two conditions
$$
\hamiltonian(z_2)-\hamiltonian(z_1) = \langle \overline{\nabla} \hamiltonian(z_1,z_2),z_2-z_1\rangle,\quad
\overline{\nabla} \hamiltonian(z,z) = \nabla \hamiltonian(z),\quad \forall z, z_1, z_2\in\mathbb{R}^M
$$
For a Hamiltonian problem, $\dot{z}=J\nabla \hamiltonian(z)$ it is easily seen that the method defined as
$$
  \frac{ z^{k+1}-z^k }{ \stepsize} = J\overline{\nabla} \hamiltonian(z^k,z^{k+1})
$$
will be energy preserving in the sense that $\hamiltonian(z^k)=\hamiltonian(z^0)$ for all $k>0$.
There are many choices of discrete gradients, but most of them lead to implicit schemes and therefore have the disadvantage of being computationally expensive.

Another disadvantage is that even if it makes sense to impose energy conservation for the extended autonomised system explained above for deep learning models, it is not clear what that would mean for the original problem. It remains an open problem to understand the potential for and the benefits of using energy preserving schemes for non-autonomous Hamiltonian systems in deep learning.

\subsubsection{Splitting methods and shears}\label{sec:shears}
Splitting methods are very popular time-integration methods that can be easily applied to preserve geometric structure of the underlying ODE problems, e.g. symplecticity. The idea of splitting and composition is simply to split the vector field in the sum of two (or more) vector fields, to integrate separately each of the parts and compose the corresponding flows. For example splitting a Hamiltonian vector field in the sum of two Hamiltonian vector fields and composing their flows results into a symplectic integrator. If the individual parts are easy to integrate exactly, the resulting time-integration method has often low computational cost. 
We refer to \cite{mclachlan02sm} for an overview on splitting methods. 
An ODE on $\mathbb{R}^d$ is a shear if there exist a basis of $\mathbb{R}^d$ in which the ODE takes the form
\begin{align*}
\dot{y}_i&=0,\quad i=1,\dots, k,\\
\dot{y}_i&=f_i(y_1,\dots, y_k),\quad i=k+1,\dots, d.
\end{align*}
A diffeomorphism on $\mathbb{R}^d$ is called a shear if it is the flow of a shear. Splitting vector fields into the sum of shears allows to compute their individual flows exactly simply applying the forward Euler method. 

Consider the shears  $\mathbb{R}^n \times \mathbb{R}^n\to \mathbb{R}^n\times \mathbb{R}^n$:
$$
\begin{array}{c}
(z,p)\mapsto (z+g(p),p),\\[0.2cm]
(z,p)\mapsto (z,p+f(z)).
\end{array}
$$
In the case of  autonomous, separable, Hamiltonian systems, the symplectic Euler method \eqref{sympEul} can be seen as the composition of two such maps where 
$$g(p):=h\partial_pT(p),\qquad f(z):=-h\partial_zV(z).$$
 Another popular example is the St{\"o}rmer-Verlet integrator or leapfrog integrator which is also the composition of two shears. It is possible to represent quite  complicated functions with just two shears. The ``standard map" (also known as Chirikov–Taylor map) is an area preserving, chaotic shear map much-studied in dynamics and accelerator physics. 

Shears are useful tools to construct numerical time-integration methods that preserve geometric features. As already mentioned symplecticity is one such property (if $f$ and $g$ are gradients), another is the preservation of volume and, as we will see in section~\ref{sec:invertible}, shears can be used to obtain invertible  neural networks. 

\subsection{Features evolving on Lie groups or homogeneous manifolds}\label{sec:Riemanniandata}

If the data belongs naturally to a differentiable manifold $\mathcal{M}$ of finite dimension and  $f(z,\theta)$ in  \eqref{eq1:ode} is a vector field on $\mathcal{M}$ for all $\theta \in\Theta$ %$\theta : [0, T] \to  R^{M^2+M}$ sufficiently smooth 
then $z(t)\in \mathcal{M}$. Concrete examples of data sets in this category are manifold valued images and signals. One example is  diffusion tensor imaging consisting of tensor data which at each point $(i,j)$ in space corresponds to a $3\times 3$ matrix $A_{i,j}\in\mathrm{Sym}^+(3)$  symmetric and positive definite, \cite{cook06cos}. If $m\times l$ is the number of voxels in the image then $\mathcal{M}=\mathrm{Sym}^+(3)^{m\times l}$. Another example is InSAR imaging data taking values on the unit circle $S^1$, and  where $\mathcal{M}=\left({S^1}\right)^{m\times l}$, \cite{rocca97aoo}. In both these examples neural networks, e.g. denoising autoencoders \cite{vincent10sda}, can be used to learn image denoising. The loss function \eqref{eq1:training} can take the form
$$\Loss (\Psi(x, \theta), y) =\sum_{i,j=1}^{l,m}\mathrm{d}(\Psi(x,\theta)_{i,j},y_{i,j})$$
where $x=(x_{1,1},\dots ,x_{l,m})\in\mathcal{M}$ is the noisy image $y=(y_{1,1},\dots ,y_{l,m})\in\mathcal{M}$ is the source image and  $\mathrm{d}$ is a distance function on $\mathrm{Sym}^+(3)$ or $S^1$ respectively.
For example in the case of $\mathrm{Sym}^+(3)$ with $T_A\mathrm{Sym}^+(3)=\mathrm{Sym}(3)$, a possible choice of Riemannian metric is
$$\langle X,Y\rangle_{T_A\mathcal{M}}:=\mathrm{trace}(A^{-\frac{1}{2}}XA^{-1}YA^{-\frac{1}{2}})$$
where $X$ and $Y$ are symmetric $3\times 3$ and $A$ is symmetric and positive definite, and with $\mathrm{trace}$ denoting the trace of matrices.
With this metric $\mathrm{Sym}^+(3)$ is a Riemannian symmetric space  and is geodesically complete \cite{weinmann14tvr}. %{\it (The space of symmetric positive definite matrices is a Riemannian symmetric space, it is acted upon by $GL(n)$ with the action $(g,A)\mapsto gAg^T$ the chosen metric is invariant under this action (the action is isometric with respect to the metric), $\mathrm{Sym}=GL(n)/H$ where $H$ is the isotropy of this Lie group action.  ) }

A third example concerns classification tasks where the data are Lie group valued curves for activity recognition, here $\mathcal{M}=G^m$ with $m$ the number of points where the curve is sampled and $G=\mathrm{SO}(3)$ is the group of rotations, \cite{CMU}.
%The 
A loss function can be built using for example the following distance between two sampled curves $c_1=(c_{1,1},\dots , c_{1,m})\in G^m$, $c_2=(c_{2,1},\dots , c_{2,m})\in G^m$,
$$\mathrm{d} (c_1, c_2) = \sum_{i=1}^{m-1}\left \|  \frac{ \log(c_{1,i+1}c_{1,i}^{-1}) }{ \| \log(c_{1,i+1}c_{1,i}^{-1}) \|_{\mathfrak{g}}^{\frac{1}{2} }  }
- \frac{ \log(c_{2,i+1}c_{2,i}^{-1}) }{ \| \log(c_{2,i+1}c_{2,i}^{-1}) \|_{\mathfrak{g}}^{\frac{1}{2} }  }      
\right\|_{\mathfrak{g}},$$
where $\mathfrak{g}$ is the Lie algebra of $G$, $\| \cdot \|_{\mathfrak{g}}$ is a norm deduced from an inner product on $\mathfrak{g}$, and $\log:G \rightarrow \mathfrak{g}$ denotes the matrix logarithm.
An important feature of this distance is that it is re-parametrisation invariant and taking the infimum over all (discrete) parametrisations of the second curve one obtains a well defined distance for curves independent on their parametrisation, see \cite{celledoni16sao} for details.

The ODE \eqref{eq1:ode} should in this setting be adapted to be an ODE on $\mathcal{M}$. If $\mathcal{M}$ is a $d-p$ submanifold of $\mathbb{R}^d$, $\mathcal{M}:=\{y \, | \, g(y)=0\}$, $g:\mathbb{R}^p\rightarrow \mathbb{R}^d$ then the ODE \eqref{eq1:ode} can be modified adding constraints, alternatively one could consider intrinsic manifold formulations which allow to represent the data with $d-p$ degrees of freedom instead of $d$.  The numerical integration of this ODE must then respect the manifold structure.

\subsection{Open problems}

\subsubsection{Geometric properties of Hamiltonian models}
Autonomous Hamiltonian systems and their numerical solution are by now very well understood. 
For such models, one has conservation of energy that is attractive when considering stability of the neural network map.
The same cannot, to our knowledge, be said about the non-autonomous case. One can approach this problem in different ways. One is to consider canonicity in the sense of \cite{asorey83gct} and study the geometric properties of canonical transformations via the extended autonomous Hamiltonian system. This is however not so straightforward for a number of reasons. One issue is that every level set of the extended Hamiltonian will be non-compact. Another issue is that the added conjugate momentum variable, denoted $p_t$ above, is artificial and is only used to balance the time varying energy function.

One should also consider the effect of regularisation, since one may expect that smoothing of the parameters will cause the system to behave more similarly to the autonomous problem. See subsection~\ref{subsec:regularisation} of this paper.

\subsubsection{Measure preserving models}
The most prevalent example of measure to preserve is the phase space volume. In invertible networks some attention is given to volume preserving maps, see \cite{rezendenormalizingflows,dinh2014nice} and also section~\ref{sec:invertible} of this paper.
For the ODE model this amounts to the vector field $f(t,z)$ being divergence free, and there are several ways to parametrise such vector fields. All Hamiltonian vector fields are volume preserving, but the converse is not true. Volume preserving integrators can be obtained, for example, via splitting and composition methods \cite{mclachlan02sm}.

\subsubsection{Manifold based models}
A generalisation of most of the approaches presented in this section to the manifold setting is still missing. For data evolving on manifolds the 
ODE \eqref{eq1:ode} should be adapted to be an ODE on $\mathcal{M}$. Just as an example, a gradient flow on $\mathcal{M}$ analogous to \eqref{gradsys} can be considered starting from defining a function $V:\mathcal{M}\times \Theta \rightarrow \mathbb{R}$ using the antiderivative of the activation function $\sigma$, and the Riemannian metric.
The Hamiltonian formulations of section~\ref{subsubsec:hamilton} could be also generalised to manifolds in a similar way, starting from the definition of the Hamiltonian function.

An appropriate numerical time-discretisation of the ODEs for deep learning algorithms must guarantee that also the evolution through the layers remains on the manifold so that one can make use of the Riemannian structure of $\mathcal{M}$ and obtain improved convergence of the gradient descent optimisation, see also section~\ref{subsection:LearningRiemannian}. 
The numerical time discretisation of this ODE must then respect the manifold structure and there  is a vast literature on this subject, see for example \cite[Ch. IV]{hairer2006geometric}. For numerical integration methods on Lie groups and homogeneous manifolds including symplectic and energy-preserving integrators see \cite{iserles00lgm,celledoni14ait}.

\section{Deep Learning meets Optimal Control}\label{sec:optimalcontrol}
As outlined in the introduction supervised deep learning with the ResNet architecture can be seen as solving a discretised optimal control problem. This observation has been made in \cite{haber2017stable, E2017deepode, Ciccone2018ode, Lu2018ode} with further extensions to PDEs are described in \cite{Ruthotto2019pde}.

Recall \eqref{eq1:training} 
\begin{align}
    \min_{\theta \in \Theta} \left\{ E(\theta) = \frac1N \sum_{n=1}^N \Loss_n(\Psi(x_n, \theta)) + R(\theta)\right\} \label{eq3:reduced}
\end{align}
where the neural network $\Psi(\cdot, \theta) : \mathcal X \to \mathcal X$ is either defined by a recursion like \eqref{eq1:discreterecursion} or the solution at the final time of an ODE \eqref{eq1:ode}. Another approach is to view the training as an optimisation problem over $\Theta \times \FeatureSpace^N$ where $\FeatureSpace$ is the space of the dynamics $z$, i.e.
\begin{subequations}\label{eq3:optimalcontrol}
\begin{align}
    &\min_{(\theta, z) \in \Theta \times \FeatureSpace^N} \left\{ E(\theta, z) = \frac1N \sum_{n=1}^N \Loss_n (z_n(T)) + R(\theta)\right\} \\
    \text{such that} \quad &\dot \HiddenVar_n = f(\HiddenVar_n, \theta(t)), \quad \HiddenVar_n(0) = x_n, \quad n = 1, \dots, N. \label{eq3:ode}
\end{align}
\end{subequations}
In machine learning the reduced formulation \eqref{eq3:reduced} is much more common than \eqref{eq3:optimalcontrol}. 

\subsection{Training algorithms}

The discrete or continuous optimal control problem can be solved in multiple ways, some of which we will discuss next.

\subsubsection{Derivative-based algorithms}
The advantage of the reduced problem formulation \eqref{eq3:reduced} is that it is a smooth and unconstrained optimisation problem such that if $\Theta$ is a Hilbert space, derivative-based algorithms such as "gradient" descent are applicable. Due to the nonlinearity of $\Psi$ we can at most expect to converge to a critical point $E^\prime(\theta) = 0$ with a derivative-based algorithm. The most basic algorithm to train neural networks is stochastic gradient descent (SGD)~\cite{robbins1951stochastic}. Given an initial estimate $\theta^0$, SGD consists of iterating two simple steps. First, sample a set of data points $\mathcal S^j \subset \{1, \dots, N\}$ and then iterate
\begin{align}
    \theta^{j+1} = \theta^j - \tau^j \frac{1}{|\mathcal S^j|} \sum_{n \in \mathcal S^j} (L_n \circ \Psi(x_n, \cdot))^\prime(\theta^j).
\end{align}
Other first-order algorithms used for deep learning are the popular Adam \cite{kingma2014adam} but also the Gauss--Newton method has been used \cite{haber2017stable}. A discussion on the convergence of SGD is out of the scope of this paper. Instead we focus on how to compute the derivatives $(L_n \circ \Psi(x_n, \cdot))^\prime(\theta^j)$ in the continuous model which is the central building block for all first-order methods. This following theorem is very common and the main idea dates back to Pontryagin \cite{pontryagin1987mathematical}. This formulation is inspired by \cite[Lemma 2.47]{Bonnans2019}.

\begin{theorem} \label{th3:bonnans}
Assume that $f$ and $L_n$ are of class $C^1$ and that $f$ is Lipschitz with respect to $z$.
Let $z_n \in W^{1, \infty}([0, T], \mathbb R^M)$ be the solution of the ODE \eqref{eq3:ode} with initial condition $x_n$ and $p_n$ the solution of the adjoint equation
\begin{align}
    \dot p_n = - \partial_z f(z_n(t), \theta(t))^T p_n, \quad  p_n(T) = L_n^\prime(z_n(T)). \label{eq3:adjoint}
\end{align}

Then the Fr\'echet derivative of 
$A := L_n \circ \Psi(x_n, \cdot) : L^\infty := L^\infty([0, T], \mathbb R^{M^2 + M}) \to \mathbb R$ at $\theta \in L^\infty$ is the linear map 
$B := A^\prime(\theta) : L^\infty \to \mathbb R$, 
\begin{align}
    Bh = \int_0^T \langle \partial_{\theta} f(z_n(t), \theta(t))^T p_n(t), h(t)\rangle \dint t . \label{eq3:frechetderivative}
\end{align}
\end{theorem}

For finite dimensional $\theta$ a similar theorem can be proven which dates back to Gr\"onwall in 1919, see \cite{Gronwall1919ode, Hairer1993ode1}. Defining neural networks via ODEs and computing gradients via continuous formulas similar to Theorem \ref{th3:bonnans} has been first proposed in \cite{chen2018neural} with extensions in \cite{Gholami2019anode} and \cite{Dupont2019anode}. In the deep learning community this is being referred to as \emph{Neural ODEs}.

Similar to Theorem \ref{th3:bonnans} a discrete version can be derived when the ODE \eqref{eq1:ode} is discretised with a Runge--Kutta method is given in \cite{Benning2019ode}, see also \cite{hager2000runge} for a related discussion about this topic. For simplicity we just state the special case of the explicit Euler discretisation (ResNet \eqref{eq1:discreterecursion}) here.

\begin{theorem}[{\cite{Benning2019ode}}] \label{th3:discrete}
Let $z_n$ be the solution of the ResNet \eqref{eq1:discreterecursion} with initial condition $x_n$. If $p_n$ satisfies the recursion
\begin{align}
    p^{k+1}_n = p_n^k - h \partial_z f(z^k_n, \theta^k)^T p^{k+1}_n, \quad k = 0, \dots, K-1, \quad  p_n^K = L_n^\prime(z_n^K), \label{eq3:gradient}
\end{align}
then the derivative of $A := L_n \circ \Psi(x_n, \cdot)$ is given by $\partial_{\theta^k}A(\theta) = h \partial_{\theta} f(z^k_n, \theta^k)^T p_n^{k+1}$.
\end{theorem}
Some readers will spot the similarity between Theorem \ref{th3:discrete} and what is called \emph{backpropagation} in the deep learning community. This observation was already made in the 1980's, e.g.~\cite{lecun1988theoretical}.

If all functions in \eqref{eq3:frechetderivative} are discretised by constant functions on $[t^k, t^{k+1}]$, then the gradient of the discrete problem \eqref{eq3:gradient} approximates the Fr\'echet derivative of the continuous problem \eqref{eq3:frechetderivative}. In more detail, let $e^{k,j} \in L^\infty$ be supported on $[t^k, t^{k+1}]$ and $e^{k,j}(t) = e^j \in \mathbb R^{M^2 + M}$ with $e^j_i = 1$ if $j = i$ and 0 else. If we denote by $A = L_n \circ \Psi(x_n, \cdot)$ the data fit using the ODE solution \eqref{eq1:ode} and $\tilde A = L_n \circ \Psi(x_n, \cdot)$ the data fit with the ResNet \eqref{eq1:discreterecursion}, then 
\begin{align}
    A^\prime(\theta)e^{k,j}
    &= \int_{t^{k}}^{t^{k+1}} \langle \partial_{\theta} f(z_n(t), \theta(t))^T p_n(t), e^{k,j}(t)\rangle  \dint t \\
    &\approx h \langle \partial_{\theta} f(z_n(t^k), \theta(t^k))^T p_n(t^{k+1}), e^j\rangle 
    = \partial_{\theta^k_j} \tilde A(\theta) .
\end{align}
In other words, in this case of piecewise constant functions, discretise-then-optimise is the same as the optimise-then-discretise.

\subsubsection{Method of successive approximation}
Another strategy to train neural networks has been recently proposed by \cite{Li2018pmpdeep} and is not based on the gradients of the reduced formulation \eqref{eq3:reduced} but on the necessary conditions of \eqref{eq3:optimalcontrol} also known as Pontryagin's maximum principle \cite{pontryagin1987mathematical} instead. Given an initial estimate of $\theta^0$, each iteration of the \emph{method of successive approximation}(MSA) has three simple steps. First, solve \eqref{eq3:ode} with $\theta^j$ which we denote by $\HiddenVar^j_n$, i.e. 
\begin{align}
    \dot \HiddenVar^j_n = f(\HiddenVar^j_n, \theta^j(t)), \quad \HiddenVar^j_n(0) = x_n, \quad n = 1, \dots, N.
\end{align}
Then solve the adjoint equation \eqref{eq3:adjoint} with $\theta^j, \HiddenVar^j_n$ and denote the solution by $p_n^j$, i.e. 
\begin{align}
    \dot p_n^j = - \partial_z f(z_n^j(t), \theta^j(t))^T p_n^j, \quad  p_n^j(T) = L_n^\prime(z_n^j(T)), \quad n = 1, \dots, N.
\end{align}
The third and final step is to maximise the Hamiltonian \eqref{eq1:hamiltonian} given $\LagrangeMultiplier_n^j, z_n^j$, i.e. for any $t \in [0, T]$ the update is defined as
\begin{align}
\theta^{j+1}(t) &:= \arg\max_{\theta}
    \left\{ H(\HiddenVar_n^j(t), \LagrangeMultiplier_n^j(t), \genparam)
    =
    \frac{1}{N}\sum_{n=1}^N \langle\LagrangeMultiplier_n^j(t), f(\HiddenVar_n^j(t), \genparam)\rangle \right\}.
\end{align}
Note that this algorithm is potentially well-defined also in the non-smooth case, i.e. $f$ is not differentiable with respect to $\theta$. If $f$ is indeed smooth, $R = 0$ and $\theta^{j+1} = \theta^j$, then
\begin{align}
    \frac{1}{N}\sum_{n=1}^N \partial_\theta f(\HiddenVar_n^j(t), \genparam^{j}(t))^T \LagrangeMultiplier_n^j(t) = 0.
\end{align}
and the Fr\'echet derivative of \eqref{eq3:reduced} vanishes. Analysis, extensions and numerical examples of the MSA are presented in \cite{Li2018pmpdeep}.

\subsection{Regularisation} \label{subsec:regularisation}
The training of a neural network \eqref{eq1:training} may include explicit regularisation $R$ and several different regularisers have been proposed in the literature, e.g. \cite{Benning2019ode,thorpe2018deep,haber2017stable,Ng2004, Ranzato2007}, which we want to discuss in this section. Before we dive into the specifics of these regularisers, we would like to answer the question if regularisation is necessary for training neural networks. 

\begin{example}
In order to shed some light on this, we consider the most trivial example which is taken from \cite{thorpe2018deep}. To this end we use ResNet \eqref{eq1:discreterecursion} and let $N = 1$, $K=1$, $M=1$, $L_1(z) = (z - 1)^2, x_1 = 0$ and $\sigma = \tanh$. When no regulariser is present, $R=0$, the training problem \eqref{eq1:training} simplifies to
\begin{align}
    \min_{A \in \mathbb R, b \in \mathbb R} \left\{ E(A, b) = (\tanh(b) - 1)^2 \right\}. \label{eq3:trivial}
\end{align}
Since $\tanh(\mathbb R) = (-1,1)$, there are two possible problems here. First, \eqref{eq3:trivial} does not have a solution, so the task of training does not really make sense. Second, even if we ignore the first problem and just apply a descent algorithm on \eqref{eq3:trivial}, then we encounter another problem: minimising sequences are not bounded. For example, let $A^j := 0, b^j := j$, then $\lim_{j\to \infty} E(A^j, b^j) = 0$ but $(A^j, b^j)_{j \in \mathbb N}$ is unbounded and does not even contain a convergent subsequence. Thus, we cannot expect our training algorithm to converge.
\end{example}

The key problem in the previous example was that the range of the neural network $\Psi(x_n, \cdot)$ was not closed. The non-closedness of the range is a characterisation of ill-posedness for linear inverse problems in infinite dimensions, see e.g. \cite[Theorem 3.7]{Clason2020}. While this may never be the case for finite-dimensional linear inverse problems, the non-linearity in \eqref{eq3:trivial} results in exactly this property.

In order to overcome this problem, we can either pose constraints on the data fidelity (network architecture, link function etc) or we cure the ill-posedness by regularisation as is classically done when considering ill-posed inverse problems, see e.g. \cite{Engl1996, Ito2014book}. To overcome the problem in \eqref{eq3:trivial} it is sufficient to add a regulariser $R$ which is \textit{coercive}, meaning that for any sequence of parameters $(\theta^j)_{j \in \mathbb N}$ it holds that
\begin{align}
    \lim_{j \to \infty} \|\theta^j\| = \infty \quad \Rightarrow \quad \lim_{j \to \infty} R(\theta^j) = \infty .
\end{align}
This condition implies that minimising sequences, which are sequences $(\theta^j)_{j \in \mathbb N}$ such that $\lim_{j \to \infty} E(\theta^j) = \inf_{\theta \in \Theta} E(\theta)$, are bounded. In reflexive Banach spaces, this it sufficient to guarantee at least a convergent subsequence. For more information on regularisation of non-linear ill-posed inverse problems we refer to classical textbooks, e.g. \cite{Engl1996, Ito2014book}.

The remainder of this section is dedicated to discuss a couple of specific choices for the regulariser $R$. In finite dimensions, due to the equivalence of all norms, any norm is coercive. In addition, the regulariser may impose additional properties on the estimated parameters. By now it is standard to use the 1-norm $\|\theta\|_1 = \sum_i |\theta_i|$ or the squared 2-norm $\|\theta\|_2^2 = \sum_i |\theta_i|^2$ for regularisation in deep learning to promote solutions which are sparse or have small coefficients respectively, see e.g. \cite{Ng2004, Ranzato2007}. The interpretation of a deep neural network as a process that changes with time naturally calls for other norms. For instance, the next section relies on the squared $H^1$-norm as a regularisation, i.e. for $\theta : [0, T] \to \mathbb R^M$ it is defined as
\begin{align}
    \|\theta\|_{H^1}^2 = \|\theta(0)\|^2 + \int_0^T \|\partial_t \theta(t)\|^2 \dint t.
\end{align}
This regularisation and its discrete counterpart will promote solutions which are smoothly varying across the layers and were used in~\cite{haber2017stable, thorpe2018deep}.

Finally, the connection of deep neural networks to discretised ODEs motivate other non-standard regularisers, too. To this end we consider the ResNet with time varying discretisation
\begin{align}
    z^{k+1} = z^k + h^k \sigma(A^k z^k + b^k)
\end{align}
and extend the parameters to $\theta = (A^k, b^k, h^k)_{k=0}^{K-1}$. Then it is natural to ensure that the time steps $h := (h^k)_{k=0}^{K-1}$ are nonnegative and sum to $T$, more precisely we regularise the time steps with $R : \mathbb R^K \to \mathbb R_\infty$,
\begin{align}
    R(h) = \begin{cases} 0, & \text{if} \quad  h^k \geq 0 \quad k = 0, \ldots, K-1, \quad  \sum_{k=0}^{K-1} h^k = T \\ \infty , & \text{else} \end{cases}.
\end{align}
Note that $R$ is nonsmooth and convex and since its proximal operator can be computed efficiently~\cite{Duchi2008} proximal algorithms can be efficiently employed. This regulariser has been used for deep learning in~\cite{Benning2019ode}.

\subsection{Deep limits} 
Let $\theta^{(K)}$ denote a minimiser of \eqref{eq1:training} with a ResNet \eqref{eq1:discreterecursion} with $K$ layers and by $\theta^\infty$ a minimiser of \eqref{eq1:training} with the ODE constraint \eqref{eq1:ode}. In what sense and under which conditions do discrete solutions $\theta^{(K)}$ converge when the number of layers $K$ tends to infinity? If these converge, do they converge to a solution of the optimal control problem~$\theta^\infty$?

In order to answer these questions one needs a topology in which we can compare $\theta^{(K)} \in (\mathbb R^{M^2+M})^K$ and $\theta^\infty : [0, T] \to \mathbb R^{M^2+M}$. 
With the discrete measure $\mu^{(K)} = \frac{1}{K} \sum_{k=0}^{K-1} \delta_{kT/K}$ we make the identification $(\mathbb R^{M^2+M})^K \cong L^2(\mu^{(K)}, [0, T], \mathbb R^{M^2+M}) =: L^2(\mu^{(K)})$ where by abusing notation, we associate the discrete object $\theta^{(K)}$ with the piecewise constant function $\theta^{(K)} : [0, T] \to \mathbb R^{M^2+M}$ with $\theta^{(K)}(t) = \theta^{(K)}_k$ if $t \in [t^k, t^{k+1})$. It turns out that $H^1 := H^1([0, T], \mathbb R^{M^2+M})$ is a suitable solution space for the optimal control problem, so that $L^2 := L^2([0, T], \mathbb R^{M^2+M})$ is a natural candidate for the convergence of $\theta^{(K)}$ to $\theta^\infty$ since both $L^2(\mu^{(K)})$ and $H^1$ can be embedded into $L^2$.

The following theorem is a special case of Theorem 2.1 in \cite{thorpe2018deep} which answers the question of deep limits for the ResNet \eqref{eq1:discreterecursion2}. Its proof relies on the equivalence of convergence in $L^2$ and a certain transport metric \cite{Garcia2016}, see \cite{thorpe2018deep} for more details.
\begin{theorem}
Let $E^{(K)} : L^2(\mu^{(K)}) \to \mathbb R$
\begin{align}
    E^{(K)}(\theta) = \frac1N \sum_{n=1}^N \Loss_n(\Psi(x_n, \theta)) + \lambda \left(\|\theta(0)\|^2 + K \sum_{k=0}^{K-1} \|\theta(t^{k+1}) - \theta(t^k)\|^2\right)
\end{align}
with $\Psi$ being the discrete ResNet \eqref{eq1:discreterecursion2} and $E^\infty : H^1 \to \mathbb R$
\begin{align}
    E^\infty(\theta) = \frac1N \sum_{n=1}^N \Loss_n(\Psi(x_n, \theta)) + \lambda \left(\|\theta(0)\|^2 + \int_0^T \|\partial_t \theta(t)\|^2 \dint t\right)
\end{align}
with $\Psi$ given by the ODE \eqref{eq1:ode}. Let $\sigma$ be Lipschitz continuous with $\sigma(0) = 0$ and $L_n$ be continuous and nonnegative for all $n = 1, \ldots, N$. If $\lambda > 0$, then \begin{enumerate}
    \item minimisers of $E^{(K)}$ and $E^\infty$ exist for all $K \in \mathbb N$,
    \item minimal values converge, i.e. $\lim_{K\to \infty} \min_{L^2(\mu^{(K)})} E^{(K)} = \min_{H^1} E^\infty$, and
    \item any sequence of minimisers of $E^{(K)}$, $\{\theta^{(K)}\}_{K \in \mathbb N} \subset L^2$, is relatively compact, and any limit point of $\{\theta^{(K)}\}_{K\in\mathbb N}$ is a minimiser of $E^\infty$.
\end{enumerate}
\end{theorem}

\subsection{Open problems}
Connecting deep learning to optimal control has opened up new avenues to advance the field of deep learning. In this section we discussed algorithms motivated by this connection which are based on derivatives and necessary optimality conditions. We discussed the need and potential for variational regularisation of deep learning and understanding the behaviour of deep neural networks as we increase the number of layers. All of these routes have natural extensions which will pave the way for better understanding of deep learning and even more powerful tools. 

\subsubsection{Algorithms with builtin errors}
Using ODEs as a network architecture and computing gradients via the adjoint, i.e. Theorem \ref{th3:bonnans}, is theoretically appealing. However, practically both the forward and the adjoint ODE have to be solved numerically which induces errors into the gradient. When using off-the-shelf first-order methods like SGD then they assume that the gradients are computed exactly which may either hinder performance or require prohibitively accurate computations, see for instance discussion in \cite{Gholami2019anode}. That being said, state-of-the-art algorithms like SGD and Adam are stochastic and the update are not guaranteed to decrease the objective so the impact of discretisation errors is not clear. Moreover, since the numerical solutions of the ODEs can be computed to any given tolerance, this naturally poses the question how to use such knowledge and control over the accuracy in optimisation algorithms. Some algorithms have been extended to include such errors, see for instance \cite[Chapter 8]{Conn2000} and \cite{Xie2020bfgs}.

\subsubsection{Algorithms without gradients}
The MSA and its extended version have been proposed for deep learning \cite{Li2018pmpdeep} but potentially more development is needed to fully exploit this direction including stochastic updates with respect to the data points and efficient maximisation of the Hamiltonian.

The MSA exploits the structure of deep learning only to the point of optimal control \eqref{eq3:optimalcontrol} but is generic in terms of the architecture, e.g. the choice of $f$. For discretised system a more tailored algorithm has been proposed in \cite{Taylor2016admm} but without any theoretical guarantees on its convergence.
    
\subsubsection{Architectures, rates and topologies for deep limits}
The question if the ResNet converges with increasing number of layers was satisfactorily answered in \cite{Thomas2018} but these results still leave a number of questions unanswered. First, do other architectures also have deep limits? The most likely candidates here are discretisations of ODEs. Second, are these results tightly linked to $H^1$-regularisaton or can they be extended to other topologies, e.g. the one induced by the total variation? Third, are there convergence rates for these limits? 

Another work on the convergence of discretised ODEs with finer discretisations is~\cite{hager2000runge}. In particular this work not only proves convergence but also convergence rates. It utilises assumptions on smoothness and coercivity of certain Hessians but it is not clear if these are met in a deep learning context. It would be interesting to verify or falsify the assumptions for relevant learning problems. In case these are not met, a different architecture or regularisation might be a way out.

\section{Invertible neural networks and normalising flows}\label{sec:invertible}
In the previous chapters, we have viewed certain neural networks as discretisations of continuous systems. In the following, we will return to the classical view of discrete networks, which are additionally endowed with a certain structure. \emph{Invertible neural networks}, i.e. bijective neural networks, are such an example. They have been an emerging topic in deep learning over the last few years. They are naturally connected to neural networks that are inspired by ordinary differential equations, as flows of ODEs are themselves invertible.
Two of the main applications of invertible neural networks lie in memory-efficient backpropagation as well as generative modeling with density estimation. For simplicity, in the following we will use the abbreviation $f_k=f^k(\cdot, \theta^k)$ when appropriate. 
Much like in general, invertible networks are typically parametrised as a function composition $\Psi=f_K \circ \dots \circ f_1$,  with the additional constraint that each layer $f_k$ is a bijective function. The inverse can then simply be computed via $\Psi^{-1}={f_1}^{-1} \circ \dots \circ {f_K}^{-1}$. A main restriction this imposes on the architecture is the fact that for each layer $f^k : \FeatureSpace^k \times \GenParamSet^k \to \mathcal{X}^{k+1}$, one has $\dim ({\mathcal{X}^k})=\dim ({\mathcal{X}^{k+1}})$, where the $\mathcal{X}^k$ are the feature spaces. As a consequence, the dimension of the input space determines the dimensions of the feature spaces in a (fully) invertible neural network.

\subsection{Types of invertible layers}
Invertible networks and layers can roughly be divided into two categories: Those that are algebraically invertible and those that are inverted with a numerical approximation scheme.

\subsubsection{Coupling layers} \label{sec:coupling}
Coupling layers \cite{dinh2014nice} work by splitting a layer's input into two parts and applying a suitable transformation that is easily invertible for one of the two parts. Mathematically, for an $M$-dimensional input, the index set $I=\{1,\dots,M\}$ is partitioned into two sets. In convolutional networks, the partition is often done channel-wise, i.e. the channels are split into two sets. A different type of partitioning is \emph{invertible downsampling} (also known as a \emph{masking} or \emph{squeeze} operation \cite{dinh2014nice}). 
With $x = (x_{1},x_{2}) \in \mathbb{R}^{M_1} \times \mathbb{R}^{M_2}$ and $y = (y_1,y_2)  \in \mathbb{R}^{M_1} \times \mathbb{R}^{M_2}$, a coupling layer $g: x\mapsto y$ is defined via the mapping
\begin{equation}
    \begin{aligned}
        y_{1} &= x_{1} \\
        y_{2} &= \gamma (x_{2}, f(x_{1})), \\ 
    \end{aligned} \label{eq:forward_coupling}
\end{equation}
where $f: \R^{M_1} \to  \R^{M_2}$ and $\gamma: \R^{M_2} \times  \R^{M_1} \to \R^{M_2}$ is invertible in the first argument. Then $g^{-1}: y \mapsto x$ is given by 
\begin{equation}
    \begin{aligned}
        x_{1} &= y_{1} \\
        x_{2} &= \gamma^{-1}(y_{2}, f(y_{1})), \\ 
    \end{aligned} \label{eq:inverse_coupling}
\end{equation}
where $\gamma^{-1}$ is the inverse of $\gamma$ in its first argument (for fixed second argument). The \emph{coupling law} $\gamma$ can be as simple as $\gamma: (a,b) \mapsto a+b$, with which $f$ is called an \emph{additive coupling layer} \cite{dinh2014nice}, such that $\gamma^{-1}: (c,b) \mapsto c-b$. Another commonly-used class of coupling layers are \emph{affine coupling layers} \cite{dinh2014nice, dinh2016density}. More complex, but in theory more expressive coupling laws can be constructed from strictly monotonic splines \cite{durkan2019neural}. Note that \eqref{eq:forward_coupling} and \eqref{eq:inverse_coupling} are shear mappings (Section \ref{sec:shears}), which in the case of ODEs are used to construct e.g. symplecticity-preserving numerical solutions.

There is in principle no restriction on the function $f$. This allows for the utilisation of arbitrarily expressive sub-networks $f$ (e.g. a sequence of convolutional layers with non-linear activation functions), without rendering $g$ (algebraically) non-invertible. The numerical stability of this inversion may still pose an issue. Stability guarantees (both for the forward and the inverse mapping) can for instance be controlled via the Lipschitz constant of $f$ and $\gamma$, as shown in \cite{behrmann2020on}. This mirrors stability considerations of ODEs (see Section \ref{sec:structure_preserving_ODE_formulations}), where guarantees can be formulated as conditions on the Lipschitz constants. Note that while $f$ has to map to $\R^{M_2}$, the individual layers which comprise $f$ may change the dimensionality throughout this sub-network -- only the final layer has to transform the data to the required space $\R^{M_2}$. Since coupling layers perform learnable computations only on a part of the input, the partitioning should change throughout the network, e.g. by switching the roles of $x_1$ and $x_2$.

\subsubsection{Invertible layers through iterative schemes} \label{sec:invertible_by_numerical_schemes}
Another class of invertible layers are those that are invertible with an iterative scheme. Invertible residual networks \cite{invertibleresnet} are a special case of the commonly-used residual networks~\cite{He2016resnet}, which allow for the inversion with a simple fixed-point iteration. A residual layer can be framed as a function $g: \mathbb{R}^M \to \mathbb{R}^M$ with
\begin{equation}
    \begin{aligned}
        g(x) = x + f(x),
    \end{aligned} \label{eq:invertible_resnet_formula}
\end{equation}
where $f: \mathbb{R}^M \to \mathbb{R}^M$ is a sub-network. For fixed $x$, let $y:=x+f(x)$, such that $x= y - f(x)$. A fixed point $z^\ast$ of the function $\Phi_y: z \mapsto y-f(z)$ is thus in the preimage of $y$ under $g$, i.e. $g(z^\ast)=y$. Note that 
\begin{equation}
    \begin{aligned}
        \| \Phi_y(a) - \Phi_y(b) \| = \| f(b) - f(a) \| \leq \text{Lip}(f) \cdot \| b-a \|,
    \end{aligned}
\end{equation}
where $\text{Lip}(f)$ is the Lipschitz-constant of $g$. According to Banach's fixed-point theorem, if $\text{Lip}(f)<1$, then $\Phi_y$ is guaranteed to have a \emph{unique} fixed point, which is the inverse of $y$ under $g$, i.e. $z^\ast=g^{-1}(y)$. This inverse can be approximated to arbitrary precision via the iteration
\begin{equation}
    x^{i} = y - f(x^{i-1})
\end{equation}
for any initial value $x^0$. While most common layer types (such as dense or convolutional layers equipped with activation functions with bounded derivative) are Lipschitz, the condition $\text{Lip}(f)<1$ is not necessarily met. In \cite{invertibleresnet}, the required Lipschitz constraint is enforced by \emph{spectral normalisation}:   
Let the linear layer $A_\theta$ depend linearly on its parameters $\theta$ (e.g. convolutional layers without non-linear activation functions and biases; these are linear layers that linearly depend on their kernel). Then $A_{\tilde{\theta}}$ parametrised by 
\begin{equation}\bar{\theta}:= c \cdot \theta / \| A_\theta \|_2 \label{eq:spectral_normalisation}\end{equation}
has Lipschitz constant $\text{Lip} (A_{\bar{\theta}}) = c$. As a consequence, the layer $f(x)=\Activation (A_{\bar{\theta}}\cdot x + b)$ has Lipschitz constant $\text{Lip}(f) \leq c$ for any activation function $\Activation$ with $\text{Lip}(\Activation) \leq 1$, where $b$ is a bias vector. Thus, by updating the layers' parameters via spectral normalisation according to \eqref{eq:spectral_normalisation} after each gradient step, the required Lipschitz constraint for invertibility is guaranteed, if one chooses $c<1$. In practice, the spectral norm $\| A_\theta \|_2$ is calculated with the power method. 
To save computations, the authors in \cite{invertibleresnet} only perform a single power iteration, but re-use the estimation of the leading singular vector from the previous training step as the initial guess. While the power method (with a finite number of iterations) technically only provides a lower bound to $\| A_\theta \|_2$, the authors in \cite{invertibleresnet} find the Lipschitz constraint to still be met in practice. \\
In the case of invertible residual networks, the connection to neural networks as numerical solutions to ODEs is particularly strong. As noted in Section \ref{sec:intro_resnets_ode}, ResNets can generally be viewed as Euler discretisations of an ODE, if one views the activations, weights and biases as observations of time-dependent variables. Fittingly, in \cite{invertibleresnet} the authors originally motivate the inversion of such a ResNets by looking at the dynamics of the associated ODE backwards in time.

\subsubsection{Linear Invertible Components}\label{linearInvertible}
Above, two general approaches for constructing invertible, nonlinear layers were presented. In the following, we list a few \emph{linear}, invertible layers, which are used to increase the expressivity of invertible networks.

Coupling layers (Section \ref{sec:coupling}) work by dimension splitting, which in the context of convolutional neural networks usually consists in splitting the channels into two groups, which are processed independently from one another. This is in contrast to standard, multi-channel convolutions, where each output channel depends on all input channels. In order to overcome this limitation, the channels can subsequently be linearly combined (via an endomorphism). If the matrix of these coefficients is invertible, the automorphism created this way can be integrated as a linear layer into the invertible network framework. These invertible matrices can be parametrised in different ways:
One approach \cite{kingma2018glow} is to directly parametrise the desired matrix via a LU factorisation (with additional fixed permutation), which can then be inverted. While not discussed in the original publication, the diagonal entries themselves can be e.g. parametrised to be larger than some $\varepsilon>0$ to enforce stable invertibility. This is known as \emph{invertible} $1\times 1$ \emph{convolution}, because this 'channel mixing' can equivalently be realised as a convolution with (in 2D) 1-by-1 filters. An extension to this idea to \emph{convolutional} layers (with larger filters) is presented in \cite{pmlr-v97-hoogeboom19a}. 
A computationally particularly cheaply (and stably) invertible class of matrices are orthogonal matrices. In e.g. \cite{NIPS2019_8336}, these are parametrised as a sequence of Householder transformations. Other possible parametrisations of (special) orthogonal matrices include exponentials of skew-symmetric matrices, Cayley transforms or sequences of Givens rotations. For a discussion about the numerical optimisation of such classes of parameter matrices compare also the forthcoming Section \ref{sec:SPT} and in particular \eqref{eq:param1}.

Another type of linear, invertible layer are invertible down- and upsampling operations for image data. While classical down- and upsampling methods (such as bilinear interpolation or nearest-neighbour methods) from image processing are inherently non-invertible (as they change the dimensionality of the input), it is possible to construct \emph{invertible} up- and downsampling methods. Since the dimensionality of their output must be the same as the dimensionality of their input, decreasing (or increasing) the spatial dimensionality of an image \emph{must} be accompanied by a suitable increase (respectively decrease) in the number of channels. One such transformation for invertible downsampling is the \emph{pixel shuffle} transform \cite{dinh2014nice}, which subsamples the pixels and reorders them into new channels. The inverse of this transformation is an invertible upsampling operation. In \cite{etmann2020iunets}, this is generalised to learnable invertible up- and downsampling operations, which contain the (inverse) pixel shuffle as a special case. The general idea is to construct \emph{orthogonal} strided convolutional operators, where the kernel size matches the stride $s \in \mathbb{N}^d$, with spatial dimensionality $d$. It can be shown that by a specific reordering of an orthogonal $\tau$-by-$\tau$ matrix (where $\tau=s_1 \cdots s_d$) into a filter kernel, the resulting convolution is orthogonal. This implies that the inverse is simply given by the corresponding transposed convolution. The authors propose to parametrise the required orthogonal matrices via exponentiating skew-symmetric matrices. As the matrix exponential is a surjective mapping from the Lie algebra of skew-symmetric matrices to the special orthogonal matrices, \emph{any} special orthogonal matrix can be parametrised this way.

\subsection{Applications} 
\subsubsection{Memory-efficient backpropagation}\label{sec:memory_efficient_backprop} One possible area of application for invertible neural networks is memory-efficient backpropagation \cite{NIPS2017_6816}. Let $\HiddenFeature^{k+1}=\Layer(\HiddenFeature^{k},{\theta^k})$ be the output of a neural network's layer with nonlinear mapping $\Layer: \FeatureSpace^k \times \GenParamSet^k \rightarrow \FeatureSpace^{k+1}$, 
where $\theta^k$ are the layer's parameters. Let further $\Loss$ be the loss of the network. Then 
$$\nabla_{\theta^k} \Loss = \left( \partial_{\theta^k} \HiddenFeature^{k+1} \right)^\ast \nabla_{\HiddenFeature^{k+1}} \Loss = \left( \partial_{\theta^k} \Layer(\HiddenFeature^{k},\theta^k) \right)^\ast \nabla_{\HiddenFeature^{k+1}} \Loss$$
provides the weight-gradient necessary for the training of the network. For the calculation of this gradient, one needs both the gradient of the loss with respect to the output node (i.e. $\nabla_{\HiddenFeature^{k+1}} \Loss$), as well as the ability to calculate the derivative $\partial_{\theta^k} \Layer(\HiddenFeature^k, {\theta^k})$. Unless $\Layer$ is linear, the calculation of the derivative requires access to $\HiddenFeature^k$, meaning that $\HiddenFeature^k$ needs to be \emph{stored in memory}. This typically represents the bulk of the memory demand in training neural networks. If, however $\Layer$ is \emph{invertible}, one can simply calculate $\HiddenFeature^{k}$ from $\HiddenFeature^{k+1}$ via
$$\HiddenFeature^{k}=\Layer^{-1}(\HiddenFeature^{k+1},{\theta^k})$$ for the additional computational cost of calculating the inverse (where $f^{-1}$ is the inverse of $\Layer$ in its first argument). Thus, instead of storing activations in memory during the forward pass, intermediate activations can simply be successively reconstructed from the output layer's activation. The memory requirement of training an invertible network using this scheme is thus independent of the number of invertible layers.

\subsubsection{Invertible networks as sub-networks}
In practice, many classical applications of neural networks such as classification, regression or segmentation do not typically map between vector spaces of the same dimensionality. Hence, a bijective function that maps between these two spaces does not exist, which is why a fully invertible neural network typically cannot solve the desired task. It is however possible to use an invertible neural networks as a \emph{subnetwork} in a neural network. For example, if $\Psi: \FeatureSpace \to \FeatureSpace$ is an invertible network and $\eta: \FeatureSpace \to \LabelSpace$ is a suitable non-invertible layer, the combined network $\eta \circ \Psi: \FeatureSpace \to \LabelSpace$ can for instance be used for a classification task of predicting labels in $\LabelSpace$ from features $\FeatureSpace$. As demonstrated in \cite{jacobsen2018irevnet}, such a neural network can have competitive performance to a comparably-sized residual network \cite{He2016resnet} on ImageNet \cite{deng2009imagenet}. Adding an output layer which transforms between spaces mirrors the approach for discretised ODEs, see Section \ref{sec:intro_resnets_ode}.

The use of invertible networks as sub-networks for example allows for the utilisation of the memory efficient backpropagation for these sub-networks (cf. Section \ref{sec:memory_efficient_backprop}), while the respective gradients of non-invertible sub-networks can be calculated conventionally. 

\subsubsection{Density estimation and generative modeling}
Aside from generative adversarial networks (GANs) and variational autoencoders (VAEs), \emph{normalising flows} are another class of machine learning models that can be used to artificially generate data. While e.g. GANs are able to generate realistic-looking images \cite{karras2019style}, they lack the ability to estimate the likelihood of data under the generative model at hand. Likewise, VAEs can only estimate a lower bound of the likelihood -- the variational lower bound. This is in contrast to normalising flows, which are trained via maximum likelihood estimation. 

\begin{figure}
    \centering
    \includegraphics[width=.3\textwidth]{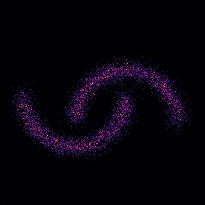}\hspace{2mm}\includegraphics[width=.3\textwidth]{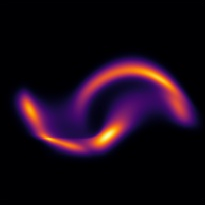}\hspace{2mm}\includegraphics[width=.3\textwidth]{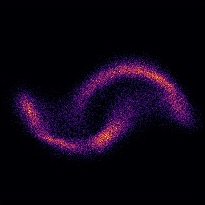}
    \caption{Left: 'Two half-moons' dataset. Middle: An approximation of the probability density function of the 'two half-moons' dataset by a normalising flow, generated by the code from \cite{chen2019residual}. Right: Points sampled from the approximated distribution.}
    \label{fig:normalising_flow}
\end{figure}

Like most generative models, normalising flows generate data from a simple base distribution (usually Gaussian) via a learned transformation. Let the random variable $z$ have probability density function $q$, for which we will write $z \sim q(z)$. For any diffeomorphism $f$, it holds that $$x:=f^{-1}(z) \sim q(z) \cdot \left| \det (\partial_z f^{-1}(z)) \right|^{-1}$$
due to the change-of-variables theorem\footnote{Here it suffices to consider invertible functions $f$ that are only \emph{locally} diffeomorphic in $x$. This is of practical relevance, because many neural networks are only locally differentiable due to the use of piecewise differentiable activation functions such as ReLU.} \cite{bogachev2007measure}. This means that for the probability density of $x$ (denoted $p$), the log-likelihood of $x$ can be expressed as
\begin{equation}
    \log p(x) = \log q(f(x)) + \log \left| \det (\partial_x f(x))  \right|. \label{eq:log_likelihood_normalising}
\end{equation}
Let $f$ be parametrised by an invertible neural network, i.e. $f(x)=\Psi(x,\theta)$ for all $x \in \FeatureSpace$. Given training data $(\Feature_n)_{n=1}^{N}$, the minimiser of
\begin{align}
    \min_{\theta \in \Theta} \left\{ E(\theta) = -\frac1N \left( \sum_{n=1}^N -\log q(\Psi(x_n,\theta)) + \log \left| \det (\partial_{x_n} \Psi (x_n,\theta))  \right| \right)\right\} . \label{eq1:normalising_flows_minimisation}
\end{align}
is a maximum likelihood estimator of the training data under the neural network. Models trained this way are called \emph{normalising flows}, because they are usually trained to convert complicated data into normal distributions. In this framework, artificial data can be generated by sampling $f^{-1}(z)$, where $z\sim q(z)$. An example of learning the 'two half-moons' dataset this way is provided in Figure \ref{fig:normalising_flow}. This case highlights that continuously transforming a normal distribution into a distribution whose probability density function has disconnected support is not possible. However, the normalising flow instead assigns low probability density to areas outside of the support, which results in a distribution that resembles the true distribution.

One of the main difficulties in designing normalising flows lies in the evaluation (respectively differentiation) of the determinant term. Note that for a neural network of the form \eqref{eq:neuralnet_general_form} it holds that $$\det (\partial_{z^0} \Psi (z^0,\theta))=\det (\partial_{z^{K-1}} z^K) \cdots \det  (\partial_{z^{0}} z^1),$$
such that the computation of the determinant for the whole network can be performed in a layer-by-layer fashion. Still, na\"{i}vely computing each determinant by the Leibniz formula yields prohibitive computation times. Depending on the types of layers used, the structure of the individial layers' Jacobian makes employing different determinant identities possible. If e.g. the Jacobian is lower triangular (as is the case with coupling layers \cite{dinh2014nice}), the determinant reduces to the product of the Jacobian diagonal entries. Invertible residual networks \cite{invertibleresnet} on the other hand induce no such structure. By using the Lipschitz constraint on $f$ in equation \eqref{eq:invertible_resnet_formula} and using the identity from \cite{withers2010log}, the authors show that in their case
$$\log \left| \det (\partial_{z^0} \Psi (z^0,\theta)) \right| = \text{trace}\left( \log \partial_{z^0} \Psi (z^0,\theta)\right)$$
holds. The matrix logarithm is then approximated as a truncated Taylor series, whereas for the evaluation of the trace, the Hutchinson estimator \cite{hutchinson1990stochastic} is employed. An extension to this concept is presented in \cite{chen2019residual}, where the truncation with a fixed number of steps is replaced by a 'Russian roulette' estimator, a type of Monte-Carlo estimator. This estimator introduces a stochastic truncation of the series, which results in an unbiased estimator of the log-likelihood \eqref{eq:log_likelihood_normalising}.\\
In the context of normalising flows, another connection to ODEs becomes apparent: As noted in \cite{chen2018neural}, a continuous version of the change-of-variables theorem may be used to formulate a continuous normalising flow, which in turn can be solved numerically via a discretisation scheme. A maximum likelihood estimator can in turn be obtained by training a Neural ODE \cite{chen2018neural}.

\subsection{Open problems}
\subsubsection{Fast inverses without coupling}
Coupling layers (Section \ref{sec:coupling}) allow both for quick forward and inverse computations. Their expressivity is however hindered by the fact that expressive, learned transformations are only applied to a part of their input, while the remaining portion of the input is not transformed. However, while this problem can be mitigated by using multiple coupling layers (with different partitions of the input), it may be desirable to construct invertible layers, where each output neuron depends on all input neurons (as is the case with 'conventional' fully-connected or convolutional layers). Invertible residual layers (Section \ref{sec:invertible_by_numerical_schemes}) on the other hand fulfill this criterion, but they rely on a numerical inversion via a fixed-point iteration, which requires multiple evaluations of the forward operation. Furthermore, in order to guarantee their invertibility, a Lipschitz constant needs to be controlled, which requires additional computation time. For practical purposes, it would be desirable to combine both methods' advantages in order to obtain both fast and expressive invertible layers.

\subsubsection{Stability guarantees}
As discussed in \cite{behrmann2020on}, controlling the Lipschitz constant of invertible layers is a conceptionally simple, but practically costly method of guaranteeing a certain stability of the inverse. Furthermore, as discussed above, they need to be controlled in order to guarantee the convergence of the iterative scheme for the inversion of residual layers. In the interest of computational efficiency, is there a computationally cheaper way to control the Lipschitz constant than via the proposed power method? Work in this direction has already been performed in \cite{chen2019residual}, as the authors experiment with Lipschitz constant corresponding to mixed norms. 

In a similar vein, are there alternative ways of guaranteeing stability other than by controlling the Lipschitz constant?

\section{Equivariant neural networks}\label{sec:equivariant}

In recent years, the use of convolutional architectures in deep neural networks \cite{lecun1998convolutional} for imaging tasks in machine learning has proven to be an extremely fruitful idea. A particularly well known example of this is the work in \cite{Krizhevsky2012}, where deep convolutional neural networks (CNNs) were used to achieve state-of-the-art performance in the ImageNet contest (a challenging image classification task), outperforming other approaches by a large margin. Another striking example of the power of CNNs is given in \cite{Ulyanov2017}, where it is shown that even an untrained CNN can be used as an effective prior for natural images. It is commonly understood that the effectiveness of CNNs in imaging tasks is in large part due to them being in some sense right for the problems at hand. CNNs combine the flexibility of neural networks (in the form of many learnable filters) with the known symmetries of images: both convolutions and pointwise nonlinearities commute with translations of the underlying domain, so that the outputs of a CNN transform in a predictable way when their inputs are translated. By constraining the search space in a principled way (in theory, fully connected networks are at least as expressive as CNNs), CNNs can make efficient use of training data to learn to perform tasks to a higher standard than fully connected networks and it is easier to interpret the action of a CNN on its inputs than it is do the same for a fully connected network.

Given the success of CNNs in machine learning, it is natural to ask whether the concept of convolutional architectures can be generalised to incorporate other symmetries into neural network architectures. One current line of research in this direction is the study of group equivariant neural networks, which has been gaining considerable traction since the work on group convolutional neural networks in \cite{Cohen2016}. A neural network can be thought of as a function taking inputs from a feature space $\FeatureSpace^1$ and returning outputs from a feature space $\FeatureSpace^2$. We will call a function $\Network : \FeatureSpace^1 \to \mathcal \FeatureSpace^2$ $G$-equivariant if there are group actions of $G$ on $\FeatureSpace^1$ and $\FeatureSpace^2$ (denoted by $T^{\FeatureSpace^1} $ and $T^{\FeatureSpace^2}$ respectively to emphasise that the group actions need not be of the same nature) such that
\begin{equation}
  \label{eq:general_equivariance}
  \Network(T^{\FeatureSpace^1}_g x) = T^{\FeatureSpace^2}_g [\Network(x)]\quad\text{for all }x\in\FeatureSpace^1, g\in G.
\end{equation}
To elucidate this definition, let us note some examples of behaviour covered by it:
\begin{itemize}
\item if $G$ acts trivially on $\FeatureSpace^2$, i.e.\! $T^{\FeatureSpace^2}_g = \id$, we recover invariance of $\Network$ to group transformations of its input, which is often a desirable property of an image classifier,
 \item whereas if $\FeatureSpace^1 = \FeatureSpace^2$ and $T^{\FeatureSpace^1} = T^{\FeatureSpace^2}$, the output of $\Network$ transforms in exactly the same way as the input does, which is a useful property to have in many image-to-image tasks such as segmentation.
\end{itemize}
If we are given two $G$-equivariant functions $\Network_1 :\FeatureSpace^1\to \FeatureSpace^2$, $\Network_2:\FeatureSpace^2\to\FeatureSpace^3$ (with the same action of $G$ on $\FeatureSpace^2$ for both functions), their composition is easily seen to be $G$-equivariant:
\[\Network_2(\Network_1(T^{\FeatureSpace^1}_gx)) = \Network_2(T^{\FeatureSpace^2}_g [\Network_1(x)]) = T^{\FeatureSpace^3}_g [\Network_2(\Network_1(x))].\]
Appealing to this result and noting the usual structure of a neural network as a composition of an alternating sequence of affine maps and nonlinearities, there is a promising way of designing $G$-equivariant neural networks: design $G$-equivariant linear maps and $G$-equivariant nonlinearities, add biases as appropriate to get affine maps from the linear maps, and compose the affine maps and nonlinearities as you would in an ordinary neural network.

As it turns out, when the group actions considered above are in fact group representations (i.e.\! they act linearly), the problem of finding and characterising $G$-equivariant linear maps reduces to the well-studied problem of finding and characterising intertwiners in representation theory. This insight has recently been used to unify existing approaches to $G$-equivariant neural networks and to show that $G$-equivariant linear maps necessarily take the form of a type of group convolution \cite{Cohen2018}. Let us describe this work in some more detail here.
\subsection{Equivariant transformations of feature maps on homogeneous spaces}
\subsubsection{Homogeneous spaces}
It has been observed \cite{Cohen2018, Kondor2018c} that there is a common setting unifying many of the existing approaches to $G$-equivariant neural networks. We are given a group $G$ which acts continuously and transitively on a domain $X$ ($X$ is a so-called homogeneous space of $G$). With this assumption we can cover the cases where $X=\R^d$ and $G = \SE(d) := \R^d\rtimes \SO(d)$ the group of rotations and translations, and where $X=S^{d-1}$ and $G= \SO(d)$, which are two commonly studied cases. Fixing a point $p\in X$ as the origin and denoting by $G_p = \{g\in G| gp = p\}$ the stabiliser of $p$, we can identify $X$ with the quotient space $G/G_p$: since $G$ acts transitively on $X$, for any $q\in X$ there is a $g\in G$ such that $gp = q$. On the other hand if $g_1 p = g_2 p$, then $g_2^{-1}g_1 p =p$, so that $g_2^{-1}g_1\in G_p$, or $g_2 G_p = g_1 G_p$. Hence, there is a well-defined bijective map $X\to G/G_p$ mapping $q$ to $gG_p$, where $g\in G$ is such that $gq = p$. Conversely, given a closed subgroup $H< G$, $G$ acts transitively on the quotient space $G/H$ by left multiplication, making it into a homogeneous space of $G$. From this reasoning, we see that the homogeneous spaces of $G$ can be identified precisely with the quotient spaces $G/H$ as $H$ ranges over closed subgroups of $G$. Henceforth, we will consider two arbitrary subgroups $H_1, H_2< G$ and the associated homogeneous spaces $G/H_1, G/H_2$.
\subsubsection{Equivariant linear maps}
Scalar-valued features on these spaces can be modelled as functions $G/H_i \to \R$ and these can be arbitrarily stacked to get feature maps $x:G/H_i\to \R^C$ that transform under the action of $G$ according to $[\pi_1(g)x](p) = x(g^{-1}p)$, but in this setting one can also consider more general features: given a representation $(V_i, \rho_i)$ of $H_i$, we can consider signals to be fields of $V_i$-valued features on $G/H_i$, which are strictly more general than the stacks of scalar-valued fields since their components are mixed under the action of $G$. This is mathematically formalised by noting that $G$ is a principal $H_i$-bundle, constructing from this the associated vector bundle $P_i$ with fiber space $V_i$, the sections of which, $\Gamma(P_i)$, are the signals of interest. Under the action of the group $G$, these signals naturally transform according to the representation of $G$ induced by $H$, $\pi_{H_i}^G$. To put this in the notation of \eqref{eq:general_equivariance}, we are taking $\FeatureSpace^1 = \Gamma(P_1), \FeatureSpace^2 = \Gamma(P_2)$ and $T^{\FeatureSpace^1}_g = \pi_{H_1}^G(g), T^{\FeatureSpace^2}_g = \pi_{H_2}^G(g)$ and we are asking what the general form is of a linear map $\Network : \FeatureSpace^1\to\FeatureSpace^2$ that satisfies  \eqref{eq:general_equivariance} in this case. There are multiple ways in which $\Gamma(P_i)$ and $\pi_{H_i}^G$ can be modelled, but for this purpose it is easiest to consider Mackey functions: we identify $x\in \Gamma(P_i)$ with $x:G\to V_i$ satisfying $x(gh) = \rho_i(h^{-1}) x(g)$ for all $h\in H_i$, in which case the induced representation is just given by $[\pi_{H_i}^G(g)x](g') = x(g^{-1}g')$. Writing $\Network$ as integration against a kernel $K:G\times G\to \Hom(V_1, V_2)$, the equivariance condition tells us that
\begin{align*}\int\limits_G K(g'^{-1}g, g'') x(g'')\dint g''&= [\pi_{H_2}^G(g')\Network(x)](g)\\
  &=\Network([\pi_{H_1}^G(g')x])(g)\\
  &=\int\limits_G K(g, g'')x(g'^{-1} g'')\dint g''.
  \end{align*}
The final expression on the right hand side can be rewritten assuming left-invariance of the measure (as can be ensured if we have a Haar measure on $G$) to give
\[\int\limits_G (K(g'^{-1}g, g'') - K(g, g'g'')) x(g'')\dint g = 0 \quad \forall x\in \Gamma(P_1),\]
which is the case if and only if
\[ K(g'^{-1}g, g'') = K(g, g' g'')\quad \forall g, g', g''\in G.\]
This final condition tells us that $K(g, g') = K(e, g^{-1} g') =:k(g^{-1} g')$, so we find that $\Network$ is equivariant and only if it is given by a convolution type operation:
\begin{equation}\Network(x)(g) = \int\limits_G k(g^{-1} g') x(g')\dint g'.
  \label{eq:convolution_condition}
\end{equation}
Since we have assumed that $\Network(x)\in \Gamma(P_2)$ is a Mackey function, we must have
\begin{align*}\int\limits_G\rho_2(h_2^{-1})k(g^{-1}g') x(g')\dint g'&=\rho_2(h_2^{-1})\Network(x)(g) = \Network(x)(gh_2) \\
  &= \int\limits_G k((gh_2)^{-1} g') x(g')\dint g'\\&= \int\limits_G k(h_2^{-1} g^{-1} g') x(g')\dint g'.
\end{align*}
Hence $\rho_2(h_2)k(g) = k(h_2g)$ for all $h_2\in H_2, g\in G$. On the other hand, $x\in \Gamma(P_1)$ is also a Mackey function, so for any $h_1\in H_1$
\begin{align*}\int\limits_G k(g^{-1} g') x(g')\dint g' &= \int\limits_G k (g^{-1}g') \rho_1(h_1)\rho_1(h_1^{-1})x(g')\dint g'\\&= \int\limits_G k(g^{-1} g') \rho_1(h_1) x(g' h_1)\dint g'\\&=\int\limits_G k(g^{-1} g' h_1^{-1}) \rho_1(h_1) x(g')\dint g'.
\end{align*}
Here we have assumed right invariance of the measure. This implies that $k(gh_1) = k(g)\rho_1(h_1)$ for $h_1\in H_1, g\in G$. Taking together the above results, the condition for equivariance of a linear map $\Network:\Gamma(P_1)\to\Gamma(P_2)$ is that we perform a convolution type operation against a kernel $k$ satisfying the linear constraints
\[k(h_2 g h_1) = \rho_2(h_2) k(g)\rho_1(h_1)\quad \forall h_1\in H_1, g\in G, h_2\in H_2.\]
These constraints can be solved once the type of features $V_i$ have been chosen (so before training time) to find a basis for the convolution kernels that give rise to equivariant linear maps, and at training time we can learn equivariant linear maps by learning the parameters of an expansion in this basis.
\subsubsection{Equivariant nonlinearities}
Having established the general form for equivariant linear maps, the question remains how to choose equivariant nonlinearities. If we are to propose a nonlinearity $\Network: \Gamma(P_1) \to \Gamma(P_1)$, we can not in general just apply a pointwise nonlinearity as in ordinary neural networks: this will only work if the chosen representation of $H_1$ does not mix components, as is the case for instance if the representation of $H_1$ is trivial (which is always the case if $H_1 =\{e\}$) or if the regular representation of $H_1$ is chosen \cite{Weiler2017}. One way to make this work in general is by having the first layer of the network be a linear layer mapping feature maps defined on the base domain $X = G/H_1$ to feature maps defined on the group $G = G/\{e\}$ and letting all feature maps after that be defined on the group \cite{Cohen2016, Bekkers2018}.  If the chosen representation does not work well with pointwise nonlinearities, another thing that can be done is to take the pointwise norm of the feature map (which is a scalar-valued feature map), pass it through a pointwise nonlinearity, and multiply this by the feature map: $\Network(x)(p) = f(\|x(p)\|) x(p)$, as is done for instance in \cite{Worrall2017, Thomas2018}. Note that $f$ could include a learnable parameter such as a bias parameter. Another option is to combine features of different types in a tensor product \cite{Kondor2018b}, which is particularly convenient when working in Fourier space: in Fourier space the convolution operation becomes a pointwise multiplication, but it is not immediately obvious how to apply equivariant nonlinearities, so other methods performing the convolution in Fourier space have to transform back to "real" space (which is computationally expensive) before applying the nonlinearity~\cite{Cohen2018a, Esteves2018}.
\subsection{A numerical demonstration of the use of equivariant neural networks}
\begin{figure}[!htb]
  \centering
\includegraphics[scale=1]{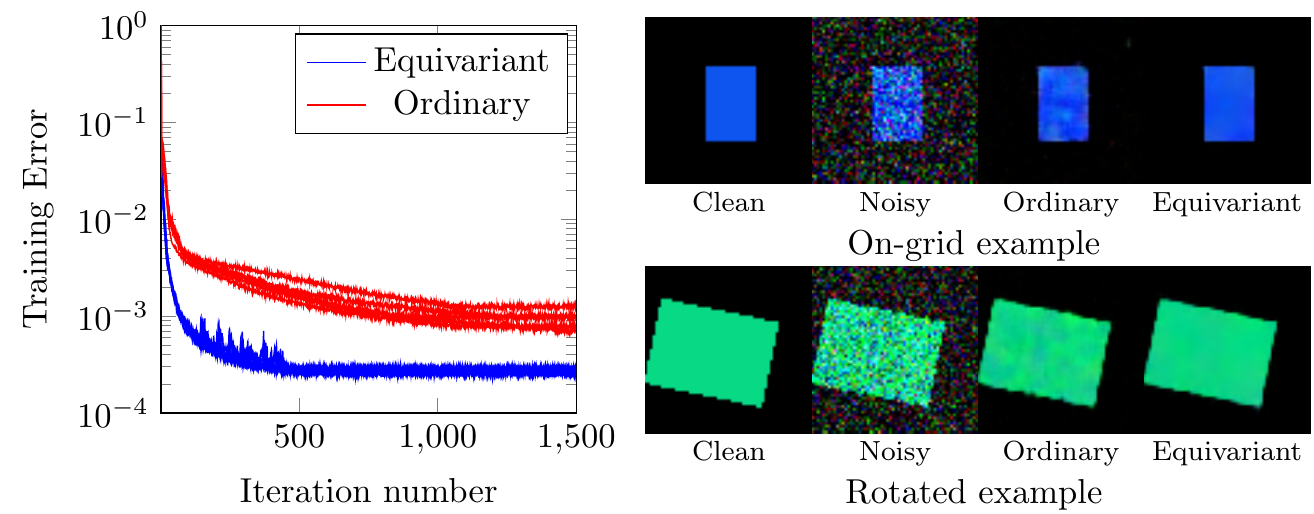}
\caption{A demonstration of the use of equivariance in a denoising task. On the left, we have plotted training errors for 5 runs of each denoiser. On the right, we have displayed the outputs of the denoisers on an example (the "on-grid" example) similar to the ones used for training and on a rotated version.}
\label{fig:equivariance_demonstration}
\end{figure}

In this section, let us consider an example that demonstrates some of the advantages that can be gained by using equivariant neural networks. We will use a supervised learning approach to learn a denoiser: we assume that we are given pairs random variables representing clean and noisy images $(x^*, x)$ and attempt to solve the following optimisation problem:
\begin{equation}
  \label{eq:toy_problem}
\min_{\Psi \in C} \frac{1}{2}\mathbf E \Big[\|\Psi(x) - x^* \|_2^2 + \lambda \| \nabla (\Psi(x)- x^*)\|_{1, \varepsilon}\Big].  
\end{equation}
Here, $C$ is a class of functions, $\lambda \geq 0$ is a small constant,  $\nabla$ is a finite difference image gradient operator and $\|\cdot\|_{1,\varepsilon}$ is a smoothed version of the 1-norm. In this specific experiment, we let the clean images $x^*$ be of size $60\times 60$, containing random rectangles with sides aligned to the grid and with random colours and we let the noisy images be generated from clean images by adding Gaussian white noise (as shown in the top right row of Figure~\ref{fig:equivariance_demonstration}). It is natural to ask for the denoiser to commute with rotations and translations, that is, for the denoiser to be equivariant with respect to rotations and translations. We compare two choices of $C$ in Problem~\eqref{eq:toy_problem}, which we will refer to as the class of ordinary and equivariant denoiser respectively. In both cases, the functions in $C$ the form of a ResNet (as defined in \eqref{eq1:discreterecursion}) preceded by a learnable lifting layer and succeeded by a learnable projection layer. The distinction between the ordinary and equivariant denoisers, is that the ordinary denoiser uses ordinary convolution operations and a pointwise ReLU nonlinearity (resulting in translation equivariance), whereas the equivariant denoiser uses the rotation and translation equivariant versions of these operations as defined in \cite{Worrall2017}. To give a fair comparison, we fix the same width and depth in both classes. In this case,  the equivariant denoiser has a number of degrees of freedom that is less than 10\% of the number of degrees of freedom of the ordinary denoiser (62994 versus 741891). The denoisers are trained by performing 1500 iterations of Adam \cite{kingma2014adam} on minibatches of size 64, decreasing the step size when validation error stagnates, and for each denoiser we perform 5 training runs. The results are shown in Figure~\ref{fig:equivariance_demonstration}: as we see in the plot of the training errors, the equivariant denoiser consistently converges in fewer iterations than the ordinary denoiser and achieves a better objective function value. Besides this, we can be confident that the equivariant denoiser will generalise to rotated examples despite not having seen them in training, whereas it is hard to say anything about how the ordinary denoiser generalises to rotated examples.
\subsection{Open problems}
\subsubsection{Sample complexity bounds}
 One of the main motivations that is given for the use of equivariant neural networks is that they should be able to use training data more efficiently than neural networks that are not designed to be equivariant. This is particularly important in applications in which training data is in short supply, as is often the case in inverse problems \cite{Arridge2019a}. In the machine learning community, the number of samples needed to estimate a given object is known as the sample complexity. There is recent work \cite{Du2018} establishing sample complexity bounds for some simple CNN models, and showing that these bounds compare favourably to the corresponding ones for fully connected networks. In a similar vein, it would be interesting to establish rigorous sample complexity bounds for equivariant neural network models that guarantee their data efficiency.
 \subsubsection{Approximation properties}
 When applying existing group equivariant neural network architectures as in the framework described above, there are a large number of choices that need to be made: the domains on which the feature maps are defined, the choices of the types of features (the representation of $H$ that is used), the choice of nonlinearity. While there is a vast amount of literature on the approximation properties for ordinary neural networks (including older works such as \cite{Cybenko1989, Hornik1991} and some more recent works that apply to CNNs \cite{Bolcskei2019, Petersen2019}), there is not yet the same theoretical guidance on how the choices of the various aspects of group equivariant neural networks can be made to guarantee that the networks are sufficiently expressive. There is some theoretical work on hypothetical equivariant neural networks and approximation results relating to them \cite{Yarotsky2018}, but it is not yet of practical use in choosing an equivariant architecture.
 \subsubsection{Approximate equivariance}
 Many of the symmetries we would like to work with in equivariant neural networks are continuous, but when we implement them in practice it is necessary to discretise them. Generally in the literature, one of two approaches is taken: the group is discretised and exact group convolutions are taken with respect to the discrete subgroup (in which case we have exact equivariance to the discrete symmetry), or the group convolutions are derived in the continuous setting and eventually discretised (so that we have approximate equivariance to the continuous symmetry). In either case, it has not been studied in detail how much of an error we make when we make these discretisations and whether there is an optimal discretisation to use. It would be of great interest to provide bounds on the equivariance error, as these could be used to decide, for example, whether a theoretically invariant classifier is actually invariant in practice.

\section{Structure-exploiting learning}\label{sec:SPT}
The training of neural networks amounts to the numerical optimisation of typically smooth, but high-dimensional and highly non-linear objectives as in \eqref{eq1:training}, and as discussed in the optimal control framework in Section \ref{sec:optimalcontrol}. The most widely used numerical method for \eqref{eq1:training} is Adam \cite{kingma2014adam}.

As before let $\Psi: \FeatureSpace\times \GenParamSet\to\FeatureSpace$ be a (deep) neural network that depends on the data and the parameters
$\theta\in\GenParamSet$. The training of $\Psi$ amounts to the optimisation over the parameters $\theta$ with respect to a loss function
$$
    \min_{\theta \in \Theta} \left\{ E(\theta) = \frac1N \sum_{n=1}^N \Loss (\eta\circ \Psi(x_n, \theta), y_n) + R(\theta)\right\},
$$
as in \eqref{eq1:training}. While the loss function $\Loss$ usually is a convex function, the dependency of the parameter $\theta$ on the network $\Psi$ is in general highly nonlinear which makes the overall optimisation problem in \eqref{eq1:training} non-convex. For what follows, let us denote by $\Loss^j(\theta)=\frac{1}{|\mathcal S^j|}\sum_{n\in \mathcal S^j} (\Loss (\eta\circ \Psi(x_n, \theta), y_n) + \frac1N R(\theta))$, for $j=0,1,\ldots$ and where $\mathcal S^j$ is a randomly chosen set of indices from the $N$ training samples. For fixed positive parameters $\alpha$ and $\beta_1,\beta_2<1$, and for appropriate initialisations for the parameters and the auxiliary moment functions $m_0$ and $v_0$, Adam amounts to the following iteration
\begin{equation}\label{eq:stochADAM}
\begin{aligned}
    m^j&=\frac{1}{(1-\beta_1^j)}\left(\beta_1 m^{j-1} + (1-\beta_1)\nabla \Loss^j(\theta^{j-1})\right)\\
    v^j&=\frac{1}{(1-\beta_2^j)}\left(\beta_2 v^{j-1} + (1-\beta_2) \nabla \Loss^j(\theta^{j-1}) \cdot \nabla \Loss^j(\theta^{j-1})\right)\\
    \theta^j&=\theta^{j-1} - \alpha \frac{m^j}{\sqrt{v^j}+\varepsilon},
\end{aligned}
\end{equation}
for a small $\varepsilon>0$ and where $\sqrt{v^j}$ is taken component-wise. Stochasticity, in the form of choosing subsets $\mathcal S^j$ randomly in every iteration, is crucial to deal with high-dimensionality of the problem.

Adam is being used for almost all of neural network training because of its easy implementation, its robustness to rescaling, its computational efficiency and small memory requirements and for its suitability for problems which are large-scale in terms of training data and parameters. On the other hand, its theoretical convergence properties do in general not guarantee convergence to a solution of \eqref{eq1:training}, cf. \cite{reddi2019convergence} where the authors propose a convergent variant in the convex setting and \cite{zaheer2018adaptive} which provides convergence guarantees in the non-convex and stochastic setting. 

The literature for optimising smooth (non-convex) objectives is, however, much richer than Adam alone. It is tightly linked to developments in convex analysis and operations research, as well as the numerical discretisation of dynamical systems, ODEs and PDEs as discussed in parts in Sections \ref{sec:ODEPDE} and \ref{sec:optimalcontrol}. Also in the context of neural network training other optimisation schemes have been investigated recently. Here, we will mainly focus on those which have some structure-preserving property such as Hamiltonian descent \cite{maddison2018hamiltonian,o2019hamiltonian} which are guaranteed to dissipate a Hamiltonian energy, optimisation approaches when the features or network parameters are elements in a Riemannian rather than Euclidean space \cite{absil08oao, udriste94cfa,li2019efficient}, and -- as a special case of the latter -- information geometry approaches that describe optimisation on statistical manifolds \cite{amari1998natural}.

\subsection{Conformal Hamiltonian systems}
The most classic approach for minimising $E(\cdot)$ over $\mathbb R^L$ is gradient descent, i.e. to seek a stationary point of $E$ by evolving
\begin{equation}\label{eq:graddescent}
\dot{\theta}=-\nabla E(\theta). 
\end{equation}
Several optimisation methods for $E$ can then be derived through different discretisations of \eqref{eq3:gradient}, with the simplest one being explicit gradient descent and its stochastic versions \cite{robbins1951stochastic}. Another route for the derivation of optimisers for $E$ is obtained by replacing the gradient flow \eqref{eq:graddescent} with a Hamiltonian flow as in \eqref{eq1:extended_ode} with dissipation, for example a conformal Hamitonian system, i.e.
\begin{equation}\label{eq:confHamiltonsys1}
\begin{aligned}
    \dot{p}&=-\nabla E(\theta)-\gamma p,\\
    \dot{\theta}&=\frac{p}{\mu},
\end{aligned}
\end{equation}
with $\gamma,\mu>0$. Rewriting the system \eqref{eq:confHamiltonsys1} in one equation gives
$$
\ddot{\theta}+\gamma \dot{\theta} = -\frac1\mu \nabla E(\theta),
$$
which is gradient descent accelerated by momentum. More generally, \eqref{eq:confHamiltonsys1} is a special case of a conformal Hamiltonian system of the form
\begin{equation}\label{eq:confHamiltoniansys2}
\begin{aligned}
    \dot{p}&=-\nabla_\theta H(p,\theta)-\gamma p\\
    \dot{\theta}&=\nabla_p H(p,\theta),
\end{aligned}
\end{equation}
for a \emph{separable} Hamiltonian function $H(\theta,p)=T(p)+E(\theta)$ as in Section \ref{subsubsec:hamilton}.   Taking $T(p)=\frac{1}{2\mu}\|p\|_2^2$ we get \eqref{eq:confHamiltonsys1}. Taking $T(p)=\sqrt{\|p\|^2+\varepsilon}$ 
we get a new optimisation approach for $E$ which is called relativistic gradient descent \cite{francca2019conformal}
\begin{equation}
\label{eq:relgraddesc}
\begin{aligned}
    \dot p &= -\nabla E(\theta)-\gamma p\\
    \dot\theta &= \frac{p}{\sqrt{ \varepsilon+\|p\|^2}}
    \end{aligned}
\end{equation}
While the gradient system \eqref{eq:graddescent} dissipates $E$, for a conformal Hamiltonian system \eqref{eq:confHamiltoniansys2} the Hamiltonian $H$ is dissipated as
$$
\frac{\ddiff}{\ddiff t} H(p,\theta) = -\frac{\gamma}{2\mu} p^T~p.
$$
For separable Hamiltonians with the kinetic energy $T$ chosen so it has a global minimiser in $0$, limit points of \eqref{eq:confHamiltoniansys2} recover stationary points of $E$. More precisely, for $H=T(p)+E(\theta)$ being separable the equilibria 
of \eqref{eq:confHamiltoniansys2} fulfill
\begin{align*}
    \gamma~p&= - \nabla E(\theta)\\
    0&=\nabla T(p).
\end{align*}
If $T$ is chosen so it has a unique global minimum in $p^*=0$ (e.g. in the case of \eqref{eq:confHamiltonsys1}), we have that $(0,\theta^*)$ is a zero of $\nabla H(p,\theta)$ if and only if $u^*$ is a zero of $\nabla E(\theta)$, i.e. $\theta^*$ is a stationary point of $E(\theta)$. Moreover, we have that $(0,\theta^*)$ is the solution of the conformal Hamiltonian system as $t\rightarrow\infty$. Then, using a numerical integration method that preserves the property of Hamiltonian energy dissipation (such a numerical method is called conformal symplectic scheme \cite{bhatt2016second}), we get an approximation of  $(\theta^*,0)$, where $(\theta^*,0)$ is an equilibrium of \eqref{eq:confHamiltoniansys2} and $\theta^*$ a stationary point of $E$ \cite{mclachlan2001conformal}. 

\begin{figure}
    \centering
    \includegraphics[height=4.5cm]{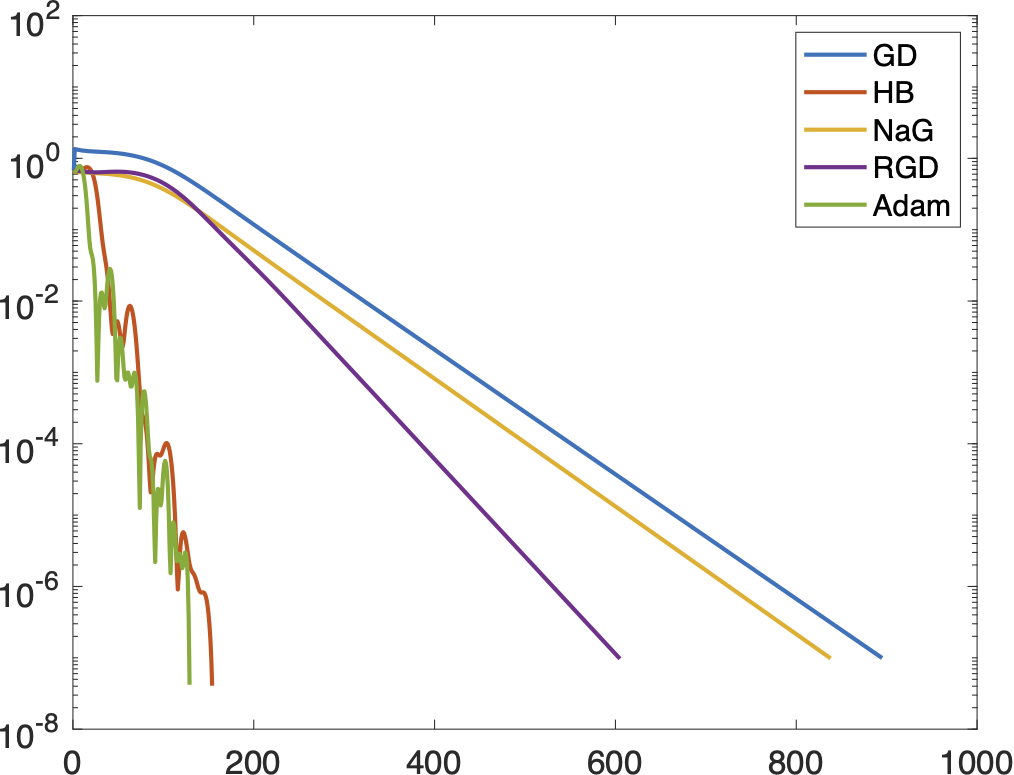}\hspace*{5mm}
    \includegraphics[height=4.5cm%, clip, trim=40px 10px 40px 20px
    ]{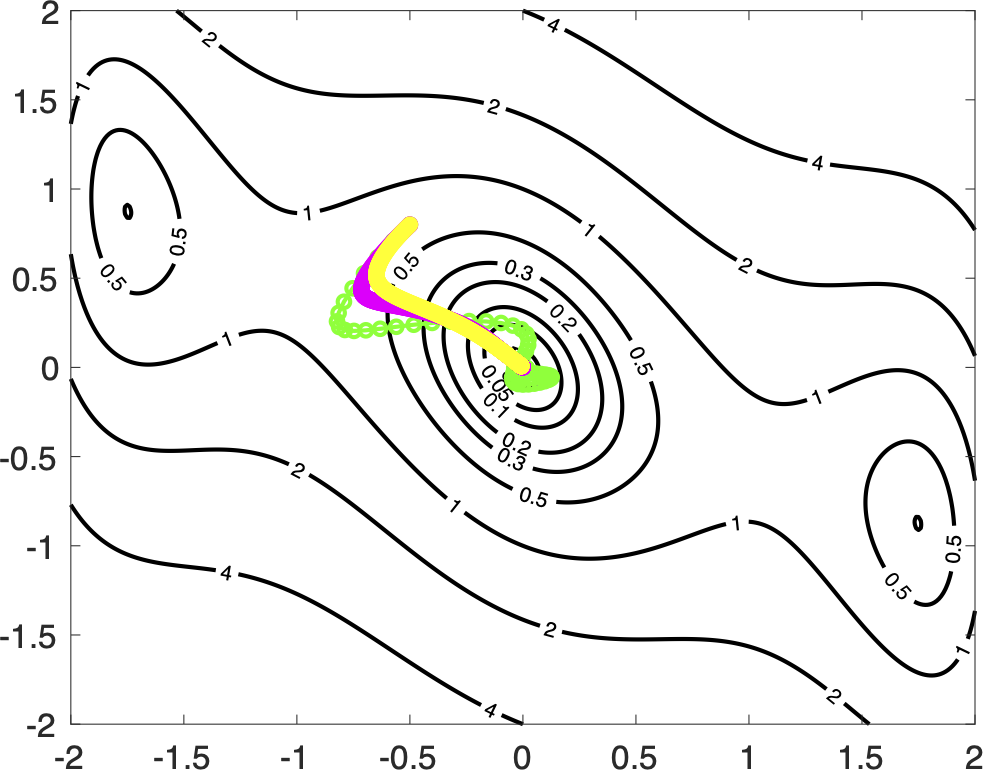}
    \caption{Optimisation of $V(x,y)=2x^2-1.05x^4 +1/6x^6 +xy+y^2$. Comparison of gradient descent, Hamiltonian descent and Adam iterations, initial guess $[-0.5,0.8]$, global minimum at $[0,0]$. Methods and parameters: Gradient descent (GD) learning rate $h=0.01$, Heavy ball (HB) $\mu=0.9$, $h=0.01$, Nesterov Accelerated Gradient (NaG) $\mu=0.012$, $h=0.01$, Relativistic Gradient Descent (RGD) \cite{francca2019conformal} $h=0.0001$, $\mu=0.9259$, Adam $\beta_1=0.9$, $\beta_2=0.999$, $\varepsilon=1e-8$, $\alpha=h=0.1$. Left: value of the loss function versus number of iterations. Right: trajectory of approach to the optimum (NaG $h=0.01$ yellow, RGD $h=0.0001$ magenta, Adam $h=0.1$ green).
    }
    \label{fig:HamiltonianDescent}
\end{figure}

In the recent work \cite{francca2019conformal} the authors take advantage of the connection between optimisation schemes for $E$ and conformal symplectic Hamiltonian schemes for $H=T+E$ for the design of new optimisation approaches for $E$ - similar to the connections we have seen before, e.g. between Adam and conformal Hamiltonian descent. See also Figure \ref{fig:HamiltonianDescent} for a comparison between different optimisation methods applied to the camelback function.

\subsection{Learning in Riemannian metric spaces}\label{subsection:LearningRiemannian}

After identifying the dissipation of an appropriate Hamiltonian as a structure worth preserving for numerical optimisation of neural networks, as discussed in the last section, explicit conditions on the parameter matrices themselves, such as orthogonality, seem to impose structure on the optimisation that can be of advantage and in particular lead to better (at least empirical) convergence rates, better generalisability and accuracy \cite{li2019efficient}. Such conditions usually pose a parameter optimisation problem within a Riemannian metric space rather than Euclidean space. Moreover, there are several important applications, cf. in particular the next subsection, where the training data and the parameters naturally lie in a Riemannian space.

While in section \ref{sec:Riemanniandata} we discussed the case when the training data and features lie on a Riemannian manifold, in this section we consider the setting when the parameters belong to a manifold. This in particular means that ordinary gradient descent a la \eqref{eq:graddescent}, i.e. with the gradient $\nabla$ being the ordinary gradient in $\mathbb R^L$, does in general not describe steepest descent of $E$ in such a Riemannian metric space. Instead descent with respect to a metric-induced gradient needs to be considered \cite{ambrosio2008gradient}. In what follows, we consider some representative examples of how Riemannian optimisation could arise in the context of deep learning.

\subsubsection{Network parameters evolving on manifolds}

Assuming that the parameters to be learned evolve on a manifold arises when additional structure or constraints are imposed on the parameters. This is done to improve stability of the training algorithm and of the trained model. Pioneering work advocating the use of Riemannian gradient and orthogonality constraints can be found in \cite{amari98ngw,amari96anl,cardoso92eas} and there is an extensive follow up literature. Examples of this procedure in the context of deep learning have recently appeared in \cite{li20ero} and earlier in the context of CNNs in \cite{cho17rat,becigneul19rao}. It is crucial to implement efficiently the Riemannian gradient descent making good use of the tensor structure of the data and of the layers of the neural network, avoiding undesired increase in computational complexity.
One way to proceed is to introduce an evolution equation for $\theta$ and replace \eqref{eq1:ode} by the extended system of ODEs
\begin{align}
\dot z &=f(z,\theta ), \label{eq:param0}\\
\dot \theta &= g( \theta).\label{eq:param1}
\end{align}
Of particular importance is the case where the second equation evolves on a Lie group $G$ or on a homogeneous manifold $\mathcal{M}=G/H$.\footnote{Here $H$ is a closed Lie subgroup of $G$, and $\mathcal{M}:=G/H$ the quotient with the manifold structure turning  $\pi: G \rightarrow G/H$  $\pi: g \mapsto gH$ into a submersion. Then $\mathcal{M}$ becomes a homogeneous space for $G$ with respect to the transitive Lie group action induced by left multiplication. }
In \cite{amari96anl} $G$ is simply the general linear group, while orthogonality constraints have been adopted in \cite{hyvarinen00ica} in the context of independent component analysis. Other concrete examples that we have in mind include the affine group, the special Euclidean group $\mathrm{SE}(n)$, the special orthogonal group $\mathrm{SO}(n)$, the Stiefel manifold $\mathrm{V}_{n,p}=\mathrm{SO}(n)/\mathrm{SO}(n-p)$ and the Grassmann manifold $\mathrm{G}_{n,p}=\mathrm{SO}(n)/(\mathrm{SO}(n-p)\times \mathrm{SO}(p))$ including the $n$-sphere.

For matrix Lie groups $G$ (and homogeneous manifolds) \eqref{eq:param1} can simply take the form of a linear matrix differential equation
$$\dot \theta = M( t)\, \theta, \qquad \theta \in \mathcal{M}$$
and $M( t)\in \mathfrak{g}$, and the optimisation can be performed in terms of the variable $M$. To preserve the manifold structure one can consider a discretisations of this equation of the form
$$\theta_{k+1}=\exp (h\,M_k)\theta_k,$$
where $\exp:\mathfrak{g}\rightarrow G$ is the matrix exponential. This discretisation can be seen as the prototype for constructing local coordinates\footnote{Otherwise called retraction maps \cite{absil08oao}.} on the manifold of parameters. Alternatively any approximation of the exponential map $\phi\approx \exp$ (e.g. by a rational approximant) preserving the property $\phi:\mathfrak{g}\rightarrow G$, can be used to construct local parametrisations of the manifold\footnote{An example for the case of $\mathrm{SO}(n)$ and $\mathrm{Sp}(n)$ is the Cayley transformation, used in the context of invertible networks in section
~\ref{linearInvertible}}. On homogeneous manifolds such as $\mathrm{V}_{n,p}$ and $\mathrm{G}_{n,p}$ additional structure on $M(t)$ can be exploited to reduce the computational cost of matrix exponentials and similar mappings \cite{celledoni04nlb}.

After time discretisation, this approach guarantees that the parameters at each layer of the network belong to manifolds which are all naturally equipped with Riemannian metrics and some of which are compact. In particular, a metric on $G$ which is $H$-right invariant (i.e. the right multiplication $R_h$ with $h\in H$ is an isometry)  descends to a Riemannian metric on $\mathcal{M}=G/H$, \cite{gallot04rg}. 
Gradient descent techniques to train the network should then exploit the Riemannian structure \cite{amari98ngw, li20ero}.
A sufficient condition for convergence of Riemannian gradient descent is that the manifolds are geodesically complete, \cite{weinmann14tvr,udriste94cfa}. All Riemannian homogeneous manifolds as well as compact Riemannian manifolds are geodesically complete
\cite[IV.4]{kobayashi96fod1}.

\subsubsection{Information Geometry}\label{sec:informationgeo} 

A special case of the Riemannian structure discussed in the previous paragraph arises when taking into account the inherent statistical properties of the underlying unknown distributions of training pairs and, connected to this, statistical properties of the network parameters $\theta$.

Treating the parameters $\theta$ as probability distributions, they can be modelled as elements on a statistical manifold with an appropriate metric. A statistical manifold is a Riemannian manifold whose points correspond to probability distributions. Gradient descent on statistical manifolds is studied in information geometry. Here, the so-called natural gradient is the proposed notion for gradient on a statistical manifold \cite{amari98ngw}. For an $L$-dimensional parameter space, equipped with a Riemannian metric tensor $G=G(\theta)=(g_{ij}(\theta))\in \mathbb R^{L\times L}$ that depends on $\theta$, the natural gradient of $E(\theta)$ is defined as
$$
\tilde\nabla E(\theta) = G^{-1}(\theta) \nabla E(\theta),
$$
where $\nabla E(\theta)$ denotes the ordinary gradient of $E$ in $\mathbb R^L$. 
The natural metric considered in this context is the Fisher information, with $G$ being the Fisher information matrix of the parameters $\theta$. Natural gradient descent then reads
\begin{equation}\label{eq:ngd}
    \dot{\theta}=-G^{-1}(\theta) \nabla E(\theta).
\end{equation}
In the case of $E(\theta)=\frac{1}{2N}\sum_{n=1}^N \|\Psi(x_n,\theta)-y_n\|^2$ a squared error loss and $G$ being the Fisher information matrix, we have
$$
G(\theta)=\frac1N\sum_{n=1}^N J_n^TJ_n,
$$
where $J_n$ is the Jacobian of $\Psi(x_n,\theta)$ with respect to $\theta$, cf. \cite{martens2014new}. Note that in this case natural gradient descent, discretised with forward-Euler, is equivalent to Gauss-Newton iteration \cite{nocedal2006numerical}. This connection between natural gradient descent and (extended) Gauss-Newton methods can be extended to more general losses as well \cite{pascanu2013revisiting}. In the continuum limit, \eqref{eq:ngd} is a gradient flow with respect to the Fisher-Rao metric, see \cite[Definition 3.1.]{modin2016geometry}.

Several works \cite{amari98ngw,amari1998natural,pascanu2013revisiting} have demonstrated advantages of using the natural gradient over the ordinary gradient in \eqref{eq:graddescent} for neural network training. The Riemannian structure seems to help against the gradient descent being trapped in flat areas of the loss function's surface \cite{yang1997natural}, and as a result the network to feature better generalisation capabilities, cf. \cite{hochreiter1997flat,keskar2016large} for different characterisations of flatness of the loss function's surface.

\subsection{Optimisation of 2-layer ReLU neural networks as Wasserstein gradient flows}\label{sec:2layerwassgrad}

The inherent gradient flow structure of training neural networks also appears when studying global optimality and generalisation properties of trained networks. Bach and Chizat \cite{chizat2018global} pick up the gradient flow formulation of neural network learning and study convergence of the learning problem \eqref{eq1:training} to a global minimiser of $E$ for 2-layer ReLU neural networks $\Phi$ in the $\infty$-width limit. In particular, their analysis makes use of the structure of a Wasserstein gradient flow formulation of \eqref{eq:graddescent} over the space of probability measures in the $\infty$-width network limit. Proving convergence to a global minimiser of $E$ also allows the study of optimal generalisation capabilities of the trained 2-layer ReLU neural network by characterising the limit (for a certain class of loss functions with exponential tails) as a max-margin classifier \cite{chizat2020implicit}. 

In \cite{chizat2018global} they consider a 2-layer neural network of the form
\begin{equation}\label{eq:2layer}
\Phi(\theta,x)=\frac1J\sum_{j=1}^J f(\theta_j,x),
\end{equation}
where
$$
f(\theta_j,x)=c_j~\max\{a_j^tx+b_j,0\},
$$
for $x\in\mathbb R^M$ and $\theta_j=(a_j,b_j,c_j)\in \mathbb R^{M+2}$. The weights $\theta$ in \eqref{eq:2layer} are learned by minimising a loss function such as \eqref{eq1:training} with optional regularisation $R(\theta)= \frac{\lambda}{L}\sum_{j=1}^L \|\theta_j\|_2^2$. In this setting, they investigate the performance (generalisation capabilities) of the learned network $\Phi$ by studying the associated gradient flow of the loss function $E$ for initial weights $\theta(0)=\theta_0 ~\sim \mathrm{i.i.d.\,}\mu_0\in P_2(\mathbb R^{n+1})$
\begin{equation}\label{eq:gradflow2layer}
\dot\theta = -m \nabla E(\theta,x).
\end{equation}
In particular, they analyse convergence of \eqref{eq:gradflow2layer} in the $\infty$-width limit, which they prove can be written as a Wasserstein gradient flow. More precisely, they parametrise the network $\Phi$ with probability measures $\mu\in P_2(\mathbb R^{d+2})$ as
$$
\Phi(\mu,x) = \int \Phi(\theta,x) \dint\mu(\theta),
$$
and the associated loss function as
\begin{equation}\label{eq:lossprobparam}
E(\mu) = \frac1N \sum_{n=1}^N \Loss (\Phi(\mu,x_n),y_n) + \lambda \int \|\theta\|_2^2 \dint\mu(\theta).
\end{equation}
In this setting they prove the following results.
\begin{theorem}[\cite{chizat2018global}] Assume that
$$
\mathrm{spt}(\mu_0)\subset \{|c|^2 = \|a\|_2^2 + |b|^2\}.
$$
As $L\rightarrow\infty$, $\mu_{t,L} = \frac1L \sum_{j=1}^L \delta_{\theta_j}(t)$ converges in $P_2(\mathbb R^{d+1})$ to $\mu_t$, the unique Wasserstein gradient flow of $E$ in \eqref{eq:lossprobparam} starting in $\mu_0$.

Moreover, assuming $\mu_0$ is `diverse' enough (cf. \cite{chizat2018global} for details). If $\mu_t$ converges to $\mu_\infty$ in $P_2(\mathbb R^{d+1})$, then $\mu_\infty$ is a global minimiser of $E$.
\end{theorem}

\subsection{Open problems}

\subsubsection{Port-Hamiltonian optimisation methods}
Generally, investigating new optimisation methods for $E$ by considering different instances of Hamiltonian descent is a promising research direction. Here, different choices of Hamiltonians or special cases of Hamiltonian systems might be advantageous for classes of loss functions and network architectures, imposing different descent dynamics. A more concrete example are numerical schemes that arise when symplectic numerical integration is applied to port-Hamiltonian systems (with different Hamiltonians), cf. \cite{van2014port} for an introduction to port-Hamiltonian systems and \cite{celledoni2017energy} for the development of structure preserving numerical integrators for port-Hamiltonian systems by using discrete gradient and splitting approaches. In \cite{massaroli2019port} the authors design a loss function and parameter optimisation dynamics of the network in such a way that the neural network itself behaves like an autonomous port-Hamiltonian system. This in turn allows a proof of convergence of the optimisation algorithm to a minimum of the loss. Taking this a step further, port-Hamiltonian systems also lend themselves to the design of locally adaptive optimisation schemes as the port-Hamiltonian structure is preserved under concatenation of port-Hamiltonians with orthogonal input-output relation.

\subsubsection{Convergence analysis for natural gradient optimisation}

While several papers seem to suggest (mostly empirically or for linear networks) that natural gradient descent helps to mitigate nuisance curvature in the neural network parameter optimisation, very little seems to have been done on this theoretically, in particular for nonlinear neural networks \cite{zhang2019fast}. In general, while for particular choices of the Riemannian metric \eqref{eq:ngd} boils down to classical optimisation schemes, such as Gauss-Newton for $G$ being the Fisher information metric, analytic results on convergence properties of natural gradient flow discretisations are generally open. Indeed, the study of \eqref{eq:ngd} for different choices of $G$ could be very interesting. For instance, if $G$ is a BFGS approximation to the Hessian of $E$, then a stochastic method was proposed and analysed in \cite{Wang2017bfgs}.

\subsubsection{Studying generalization properties of neural networks by metric gradient flows} As the example of the Wasserstein gradient flow in section \ref{sec:2layerwassgrad} has shown, metric gradient flows can serve as a useful tool for studying convergence of the network training and for the study of generalisation properties of the minimisers \cite{chizat2018global,chizat2020implicit}. It would be interesting to see if other metric gradient flows would also lend themselves to such an analysis. In connection with information geometry in section \ref{sec:informationgeo} we have seen the gradient flow with respect to the Fisher-Rao metric appearing. Could this be used to study convergence properties for other network architectures, beyond 2-layer ReLU? Are there other metrics that could be interesting to investigate for that purpose?

\section{Conclusion}

%Volume preservation as proxy for structure preservation? Similarly, learning isometries (invertibility \& $det(J)=1$)? Mass preservation in denoising, etc.?

Structure-preserving approaches to deep learning are a mean to design neural networks with guaranteed mathematical properties, as well as derive optimisation schemes that improve their training and provide means for analysing their global optimality and generalisation capabilities. In this paper we are discussing some recent examples from this emerging topic of structure-preserving deep learning. These include ODE and PDE parametrisations of neural networks for improved stability properties, an optimal control formulation for neural network training that gives rise to systematic training and regularisation procedures, invertible neural networks for large-scale deep learning, equivariant neural network architectures for the design of neural networks that preserve group transformations, and structure-preserving training of neural networks by means of Hamiltonian descent and Riemannian metric gradient flows. Together with the discussion of state-of-the-art results we also suggest a range of open problems that we identified as interesting mathematical avenues that could help to shed some more light onto the systematic design and training of deep neural networks.

\section*{Acknowledgements}
MJE would like to thank Matt Thorpe for fruitful discussions. MJE acknowledges support from the EPSRC grants EP/S026045/1 and EP/T026693/1, the Faraday Institution via EP/T007745/1, and the Leverhulme Trust fellowship ECF-2019-478. 

CE and CBS acknowledge support from the Wellcome Innovator Award RG98755.

CBS acknowledges support from the Leverhulme Trust project on ‘Breaking the non-convexity barrier’, the Philip Leverhulme Prize, the EPSRC grants EP/S026045/1 and EP/T003553/1, the EPSRC Centre Nr. EP/N014588/1, European Union Horizon 2020 research and innovation programmes under the Marie Sk\l{}odowska-Curie grant agreement No. 777826 NoMADS and No. 691070 CHiPS, the Cantab Capital Institute for the Mathematics of Information and the Alan Turing Institute. 

FS acknowledges support from the Cantab Capital Institute for the Mathematics of Information. 

EC and BO thank the SPIRIT project (No. 231632) under the Research Council of Norway FRIPRO funding scheme. 

The authors would like to thank the Isaac Newton Institute for Mathematical Sciences, Cambridge, for support and hospitality during the programmes \emph{Variational methods and effective algorithms for imaging and vision (2017)} and \emph{Geometry, compatibility and structure preservation in computational differential equations (2019)} where work on this paper was undertaken, EPSRC grant EP/K032208/1.

\bibliographystyle{plain}
\bibliography{references.bib}

\end{document}